\documentclass[accepted]{uai2025} 
                        
\usepackage[american]{babel}
\usepackage{natbib} 
\usepackage{mathtools} 
\usepackage{booktabs} 
\usepackage{tikz} 

\usepackage{microtype}
\usepackage{graphicx}
\usepackage{subfigure}
\usepackage{booktabs} 
\usepackage{multirow}
\usepackage{hyperref}
\usepackage{amsmath}
\usepackage{amssymb}
\usepackage{mathtools}
\usepackage{amsthm}
\usepackage[capitalize,noabbrev]{cleveref}
\usepackage[textsize=tiny]{todonotes}

\usepackage[utf8]{inputenc} 
\usepackage[T1]{fontenc}    
\usepackage{url}            
\usepackage{amsfonts}       
\usepackage{nicefrac}       
\usepackage{natbib}

\usepackage{comment}
\usepackage{bbm}
\usepackage{placeins} 
\usepackage{calligra} 
\DeclareMathAlphabet{\mathcalligra}{T1}{calligra}{m}{n}

\usepackage{cancel} 

\DeclareMathAlphabet{\mathsfit}{\encodingdefault}{\sfdefault}{m}{sl}
\SetMathAlphabet{\mathsfit}{bold}{\encodingdefault}{\sfdefault}{bx}{n}

\DeclareMathOperator*{\argmin}{\arg\!\min}

\DeclareMathOperator{\sign}{sign}

\usepackage{amsfonts}
\usepackage{amsmath,bm}

\theoremstyle{plain}
\newtheorem{theorem}{Theorem}[section]

\newtheorem{lemma}[theorem]{Lemma}
\newtheorem{corollary}[theorem]{Corollary}
\theoremstyle{definition}

\theoremstyle{remark}

\newcommand{\expectation}[2]{\underset{#1}{\mathbb{E}}\left[#2\right]}

\usepackage{mathrsfs}

\newcommand{\lp}{\left(}
\newcommand{\rp}{\right)}

\newcommand\numberthis{\addtocounter{equation}{1}\tag{\theequation}}

\Crefname{equation}{}{}
\Crefname{proposition_app}{Proposition}{Propositions}

\def\deltat{\hat{\delta}}

\title{Generalization Certificates for Adversarially Robust Bayesian Linear Regression}

\author[1]{Mahalakshmi Sabanayagam}
\author[3]{Russell Tsuchida\thanks{work partially done while at Data61, CSIRO}}
\author[4,5]{Cheng Soon Ong}
\author[1,2]{Debarghya Ghoshdastidar}
\affil[1]{%
    School of Computation, Information and Technology\\
    Technical University of Munich\\
    Germany
}
\affil[2]{%
    Munich Data Science Institute\\
    Technical University of Munich\\
    Germany
}
\affil[3]{%
    Monash University\\
    $^4$Data61, CSIRO\\
    $^5$College of Systems and Society\\
    Australian National University\\
    Australia
}
\affil[ ]{\texttt{sabanaya@in.tum.de, russell.tsuchida@monash.edu, chengsoon.ong@anu.edu.au, ghoshdas@in.tum.de}}
  
\begin{document}
  \DeclareFontShape{T1}{calligra}{m}{n}{<->s*[2.5]callig15}{}
\maketitle

\begin{abstract}
\vspace{-0.2cm}
  Adversarial robustness of machine learning models is critical to ensuring reliable performance under data perturbations. Recent progress has been on point estimators, and this paper considers distributional predictors. First, using the link between exponential families and Bregman divergences, we formulate an adversarial  Bregman divergence loss as an adversarial negative log-likelihood. Using the geometric properties of Bregman divergences, we  compute the adversarial perturbation for such models in closed-form. Second, under such losses, we introduce \emph{adversarially robust posteriors}, by exploiting the optimization-centric view of generalized Bayesian inference. Third, we derive the \emph{first} rigorous generalization certificates in the context of an adversarial extension of Bayesian linear regression by leveraging the PAC-Bayesian framework. Finally, experiments on real and synthetic datasets demonstrate the superior robustness of the derived adversarially robust posterior over Bayes posterior, and also validate our theoretical guarantees.
\end{abstract}
\vspace{-0.2cm}

\section{Introduction}
\label{sec:intro}
Machine learning models are vulnerable to adversarial inputs, where small, carefully crafted perturbations to the input data can significantly degrade model performance. 
These perturbations, though imperceptible to humans (e.g. in computer vision contexts), can cause models to make incorrect predictions with high confidence. 
Significant progress has been made in understanding and improving adversarial robustness of point predictors, with efforts in defense mechanisms, attack strategies, and the trade-offs between robustness and generalization \citep{szegedy2013intriguing, shhafahi2019are, li2023adersarial}. 
A key insight from this body of research is that models  susceptible to adversarial attacks often exhibit near-perfect empirical generalization --— achieving similar performance on training and test data~\citep{goodfellow2014explaining}.
This observation suggests that deriving formal guarantees is critical to understanding the interplay between generalization and adversarial robustness. 

Probabilistic models 
offer an alternative paradigm that quantifies uncertainty, a property that can detect adversarial inputs and reject uncertain predictions. 
However, despite these advantages, probabilistic models have received significantly less attention than their non-probabilistic counterparts in adversarial settings \citep{bradshaw2017adversarial, grosse2018limitations}. 
While they have been studied for robustness against outliers \citep{kim2008outlier}, label noise \citep{hernandez2011robust}, and domain shifts \citep{ovadia2019can}, their susceptibility to adversarial attacks remains largely unexplored. 
There exists no notion of adversarially robust probabilistic inference, and hence no formal generalization guarantees,
thus raising a fundamental question:

\emph{How can we define and develop adversarially robust probabilistic inference and derive generalization certificates (formal guarantees) for such models?}

In this work, we address this question by first introducing the notion of \emph{adversarially robust posteriors}. 
We achieve this by formulating an adversarial variant of the negative log-likelihood (NLL) loss --- drawing inspiration from adversarial training, one of the most effective defense strategies against adversarial attacks in standard machine learning~\citep{madry2018towards} --- and taking an optimization-centric perspective of (generalized) Bayesian inference \citep{alquier2016properties}.
In doing so, we obtain a posterior that is robust to adversarial perturbations.
We leverage the PAC-Bayesian framework, a powerful tool for deriving data-dependent generalization bounds for Bayesian predictors 
\citep{mcallester1998some,catoni2004statistical}, and derive the \emph{first} rigorous certificates for the robust posterior on linear regression.
Furthermore, we also derive PAC-Bayesian based generalization certificates for Bayes posterior obtained using the standard negative log-likelihood loss.

In summary, our main \textbf{contributions} are as follows.
\begin{table}[t]
    \centering
    \resizebox{\linewidth}{!}{
    \begin{tabular}{ccc}
    \toprule
    & Standard generalization & Adversarial generalization   \\
         & NLL $\ell$ & Adversarial NLL $\ell_{\deltat}$  \\
         & $R(\theta)=\expectation{\mathcal{D} \sim \mathcal{P}}{\ell(\theta, \mathcal{D})}$ & $R_{\deltat}(\theta)=\expectation{\mathcal{D} \sim \mathcal{P}}{\ell_{\deltat}(\theta, \mathcal{D})}$  \\
         \midrule
         \multirow{2}{*}{Bayes posterior $q$} & $\expectation{\theta \sim q}{R(\theta)}$ & $\expectation{\theta \sim q}{R_{\deltat}(\theta))}$   \\
         & \Cref{thm:std_post_std_loss} & \Cref{thm:std_post_adv_loss} \\
         \midrule
         \multirow{2}{*}{Robust posterior $q_\delta$} & $\expectation{\theta \sim q_\delta}{R(\theta)}$ & $\expectation{\theta \sim q_\delta}{R_{\deltat}(\theta)}$ \\
         & \Cref{thm:adv_post_std_loss} & Theorems~\ref{thm:adv_post_adv_loss} and \ref{thm:adv_post_adv_loss_gen}\\
         \bottomrule
    \end{tabular}} %
    \caption{Overview of our derived generalization certificates. The guarantees are derived for the standard \emph{Bayes posterior} $q$ and the novel \emph{robust posterior} $q_{\delta}$, where $\delta$ denotes the training adversarial allowance. We consider generalization to standard NLL loss (\emph{standard generalization}) as well as adversarial NLL (\emph{adversarial generalization}), with a potentially different adversarial allowance $\deltat$. \label{tab:overview}}
    \vspace{-0.4cm}
\end{table}
\begin{enumerate}[($i$), topsep=0pt,itemsep=-1ex,partopsep=1ex,parsep=1ex]
    \item In exponential families, we review a one-to-one correspondence between the adversarial negative log-likelihood loss and the class of Bregman divergences. 
    Based on this correspondence, we introduce a novel adversarial negative log-likelihood loss in \Cref{sec:adversarially-robust-posterior}. 
    This probabilistic-geometric connection allows us to solve the adversarial perturbation problem in closed-form, for all exponential families, allowing for an adversarially robust formulation of generalized linear models.
    \item We define the adversarially \emph{robust posterior} as minimizing a variational objective with this loss, thereby extending Bayesian inference to adversarially robust generalized linear models settings, in \Cref{sec:adversarially-robust-posterior}.
\item In \Cref{sec:pac_bayes_certs}, focusing on the case of a Gaussian family, we derive the PAC-Bayesian generalization certificates for the Bayes posterior and the robust posterior under two settings: 
$a)$ \emph{standard generalization}: guarantees for standard negative log-likelihood loss (\Cref{thm:std_post_std_loss,thm:std_post_adv_loss}),
and $b)$ \emph{adversarial generalization}: guarantees for adversarial negative log-likelihood loss (\Cref{thm:adv_post_std_loss,thm:adv_post_adv_loss,thm:adv_post_adv_loss_gen}). 
\Cref{tab:overview} gives an overview of our bounds. 
We experimentally validate the derived certificates in \Cref{sec:exp} showing non-trivial guarantees.
\end{enumerate}
We discuss the practical significance of the bounds, several technicalities, 
and touch on related works in \Cref{sec:related_works};  and conclude in \Cref{sec:conclusion}.

\section{Preliminaries}
\label{sec:prelim}
Our work combines elements from adversarial robustness and probabilistic inference.
We briefly outline these topics here, and establish some preliminaries.
\paragraph{Notation}
We represent the entry-wise absolute value of matrix $M$ as $|M|$, and vector Euclidean norm as $\| \cdot \|$.
We use $I_n$ for identity matrix of size $n \times n$ 
and $1_n$ for a vector of size $n$ with all ones.
We denote by $\ell\big(\theta, (x,y) \big)$ a loss evaluated on parameter $\theta$ and single $(x,y)$ data pair, and use $\mathcal{L}(\theta, \mathcal{D}) = \sum_{i=1}^n \ell\big(\theta, (x_i,y_i) \big)$ for the sum of the losses over a dataset.
The expected and empirical average errors are $R(\theta)=\expectation{(x,y) \sim \mathcal{P}}{\ell\big(\theta, (x,y)\big)}$ and $r(\theta)=\frac{1}{n} \mathcal{L}(\theta, \mathcal{D})$.

\subsection{Adversarial robustness}
We are given $n$ labeled data samples drawn i.i.d. from an unknown probability measure $\mathcal{P}$, denoted by $\mathcal{D}=\{(x_i,y_i)\}_{i=1}^n$ with $x_i \in \mathbb{X} \subseteq \mathbb{R}^d$ representing the feature vector with the corresponding label $y_i \in \mathbb{R}$. We later use $X\in\mathbb{R}^{n\times d}$ to refer to the matrix of all feature vectors, and $Y \in \mathbb{R}^n$ for the vector of all $n$ labels.
We follow the standard setting of supervised learning~\citep{bishop2007pattern,deisenroth20mathml}, where we minimize the empirical loss $\ell$ on the training set $\mathcal{D} = (X,Y) = \{x_i, y_i\}_{i=1}^n$ with respect to the parameters $\theta$ of our model,
\begin{align*}
    \theta^\ast = \argmin_{\theta} \sum_{i=1}^n \ell\big(\theta, (x_i, y_i) \big) =  \argmin_{\theta} \mathcal{L}(\theta, \mathcal{D}).
\end{align*}
For the adversarially robust setting, we consider perturbed data $\widetilde{x}_i$ but not perturbed labels. 
Following typical adversarial constructs~\citep[for example]{szegedy2013intriguing} we consider perturbations whose $\ell_2$ distance is bounded by a user-defined constant $\delta$.
\begin{align*}
    \theta^\ast = \argmin_{\theta} \sum_{i=1}^n \max\limits_{\Vert \widetilde{x}_i - x_i \Vert \leq \delta}\ell\big(\theta, (\widetilde{x}_i, y_i) \big).
\end{align*}
Here we focus on the parametric supervised setting, where a parameter $\theta$ is mapped to a prediction $f_\theta(x_i)$ on a feature $x_i$ with some parameterized function $f_\theta:\mathbb{X} \to \mathbb{F}$.

\subsection{Probabilistic inference}
\paragraph{Bayesian inference}
Of central interest in Bayesian inference is the posterior $q(\theta \mid \mathcal{D})$. 
The posterior reflects a belief of an unknown quantity of interest $\theta$ updated from a prior belief $p(\theta)$ in light of the likelihood $p(\mathcal{D} \mid \theta) = \prod_{i=1}^n p(y_i \mid x_i, \theta)$ of observations under the model.
One way to compute this update is via Bayes' rule, $q(\theta \mid \mathcal{D}) \propto p(\mathcal{D} \mid \theta) p(\theta)$.
An alternate optimization-centric perspective of Bayesian inference, introduced by \citet{csiszar1975divergence,donsker1983asymptotic},
reformulates the objective of deriving the Bayesian posterior as solving an optimization problem. 
Specifically, the Bayesian posterior distribution $q(\theta \mid \mathcal{D})$ is obtained by minimizing a variational objective,
\begin{align}
	q(\theta \mid \mathcal{D}) = \arg\min_{\rho \in \Pi} \expectation{\theta \sim \rho}{\text{-}\log p(\mathcal{D} \mid \theta)} + KL(\rho \Vert \pi), \label{eq:general_vi}
\end{align} 
where $\Pi$ is the space of all probability measures, $\pi$ is the prior on $\theta$, $\text{-}\log p(\mathcal{D} \mid \theta)$ is the negative log-likelihood loss on the data, and $KL(\rho \Vert \pi) =  \expectation{\theta \sim \rho}{\log \frac{\rho(\theta)}{\pi(\theta)}}$ is the Kullback-Leibler (KL) divergence. 

\paragraph{Generalized Bayesian inference}
Unfortunately, even under the optimization-centric view, Bayesian inference suffers from some limitations.
First, the normalizing constant and/or optimization problem can be intractable.
Second, the prior is often chosen for convenience and, particularly in large models, may not be truly calibrated to the statistician's prior beliefs.
Third, the likelihood is often also chosen for convenience, very often corresponding with losses which are not robust.
The Rule of Three (ROT)~\citep{knoblauch2022optimization} generalizes standard Bayesian inference via the optimization view, addressing the limitations above. 
Generalizing~\eqref{eq:general_vi}, the ROT replaces the negative log likelihood (NLL) with an arbitrary loss function $\mathcal{L}$, the KL divergence $KL$ with an arbitrary divergence $D$, and  the space $\Pi$ of all probability measures with a subset of all probability measures $\Lambda$,
\begin{align}
	q(\theta \mid \mathcal{D}) = \arg\min_{\rho \in \Lambda} \expectation{\theta \sim \rho}{\mathcal{L}(\theta, \mathcal{D})} + D(\rho \Vert \pi). \label{eq:rot}
\end{align} 
The ROT has axiomatic foundations and also comes with guarantees on estimation procedures. 
The variational objective balances two competing terms: $(i)$ the expected loss term, which encourages the posterior to assign a higher probability to parameters that fit the data well, and $(ii)$ the divergence term, which regularizes the posterior by penalizing deviations from the prior.

\paragraph{Gibbs Bayesian inference}
As a special case of~\eqref{eq:rot}, the Gibbs posterior addresses the problem of mis-specified and non-robust likelihoods, and also partially addresses the problem of intractability. 
The Gibbs posterior is obtained by retaining the KL divergence and space of all probability measures $\Pi$ from~\eqref{eq:general_vi} in~\eqref{eq:rot}, but using a general loss $\mathcal{L}(\theta, \mathcal{D})$ in place of the NLL $\text{-}\log p(\mathcal{D} \mid \cdot)$.
In this case, the minimizer~\eqref{eq:rot} is called 
the \emph{Gibbs posterior}, and admits a closed-form (up to the normalizing constant)~\citep[for example]{alquier2016properties,knoblauch2022optimization},
\begin{align*}
	q(\theta \mid \mathcal{D}) &= \arg\min_{\rho \in \Pi} \expectation{\theta \sim \rho}{\mathcal{L}(\theta, \mathcal{D})} + KL(\rho \Vert \pi) \\
    &=\frac{\exp\big( - \mathcal{L}( \theta, \mathcal{D}) \big) \pi(\theta) }{\int \exp\big( - \mathcal{L}( \theta', \mathcal{D}) \big) \pi(\theta') d\theta' }. \numberthis \label{eq:gibbs_posterior}
\end{align*}

\paragraph{Gaussian linear regression}
A notable special case 
arises when the loss is chosen as the negative log-likelihood $\mathcal{L}(\theta, \mathcal{D}) = - \log p(\mathcal{D} \mid \theta) = 
\sum_{i=1}^n - \log p(y_i \mid x_i, \theta)$ with isotropic Gaussian prior $\pi \sim \mathcal{N}(0, \sigma_p^2 I_d)$, as it recovers the standard Bayes posterior 
$q(\theta)= \mathcal{N}\lp \hat{\theta}, \Sigma \rp$, where $\Sigma = \frac{1}{\sigma^2}X^\top X + \frac{1}{\sigma_p^2}I_d$ and $\hat{\theta} = \frac{1}{\sigma^2}\Sigma^{-1}X^\top Y$ \citep{bishop2007pattern}.
In this work, we consider an isotropic Gaussian prior of mean zero and variance $\sigma_p^2$: $\theta \sim \mathcal{N}(0, \sigma^2_p I)$ 
and denote the 
negative log-likelihood loss on $\mathcal{D}$ as $\ell(\theta, \mathcal{D})$ which is 
$$\ell(\theta, \mathcal{D}) = \frac{n}{2}\log \lp 2\pi \sigma^2 \rp + \frac{1}{2\sigma^2} \|Y - X\theta\|^2.$$

\subsection{Linking loss functions and probability distributions}
We consider probabilistic models $p\big(y \mid f_\theta(x) \big)$ that belong to an exponential family
and associate the NLL with the notion of empirical loss $\ell$ via Bregman divergences, and vice versa.
When such losses are later subject to adversarial perturbation, this allows us to make use of the geometrical properties of the Bregman divergence (more specifically, the law of cosines) to study adversarial extensions of probabilistic models in~\Cref{lm:adv_loss_closed_form_gaussian,lm:adv_loss_closed_form}.

\paragraph{Exponential families} Generalizing Gausssian families, exponential families provide a flexible and theoretically tractable class of probability distributions~\citep[\S~6.6.3]{deisenroth20mathml}. 
For our purposes, it suffices to consider $1$ dimensional (and therefore minimal) exponential families.
Let $t:\mathbb{F} \to \mathbb{R}$ be a measurable function called a \emph{sufficient statistic}.
Let $\mu$ be a nonnegative measure, called the \emph{base measure}, defined on some appropriate sigma algebra generated by $\mathbb{F}$.
An exponential family is the set of all probability distributions (with respect to base measure $\mu$) parameterized by natural parameter $\eta$ of the form
\begin{align*}
    p(y \mid \eta) &= \exp\big( \eta t(y) - \phi(\eta) \big),
\end{align*}
where $\phi(\eta) = \log \int_{\mathbb{F}} \exp\big( \eta t(y) \big) \, \mu(dy)$ is called the log normalizing constant, such that $\phi(\eta) \in \mathbb{R}$.
We assume an extremely mild condition on exponential families, that they are \emph{regular}.
Regular means that the set of all $\eta$ such that $\phi(\eta) \in \mathbb{R}$ is an open set.
Proposition 2 of~\citet{wainwright2008graphical} then states that $\phi$ is a strictly convex function.

\paragraph{Bregman divergence}
Generalizing the \textbf{squared} Euclidean distance, Bregman divergences allow for a natural class of loss functions for use in a wide variety of supervised and unsupervised applications.
Assume $\mathbb{F}$ is a convex set and let $\phi:\mathbb{F} \to \mathbb{R}$ be a continuously differentiable and strictly convex function (so called generator).
The \emph{Bregman divergence} $d_\phi:\mathbb{F} \times \mathbb{F} \to \mathbb{R}$ generated by $\phi$ is defined by
\begin{align*}
    d_\phi(y_1, y_2) = \phi(y_1) - \phi(y_2) - \nabla \phi(y_2)^\top (y_1 - y_2),
\end{align*}
and is strictly convex in its first argument.

A link between geometric loss functions in Bregman divergences and probabilistic loss functions in NLLs is provided through the fact that (informally speaking) every NLL of an exponential family is a Bregman divergence.
More precisely, in our current context, if $p\big(y \mid f_\theta(x) \big)$ belongs to a regular exponential family with  log normalizing function $\phi$ and natural parameter $\eta = f_\theta(x)$, then by~\citet[Theorem 4]{banerjee2005clustering},
\begin{align*}
    - \log p\big(y \mid f_\theta(x) \big) &= d_\phi\big(f_\theta(x), y^\ast \big) + C(y) \numberthis \label{eq:exp_bregman_link}
\end{align*}
where $C(y)$ is an additive constant independent of $\theta$ and $x$ available in closed-form, and $y^\ast = (\nabla \phi)^{-1}(y)$ is the dual coordinate of $y$.
Note that some technical care is required in ensuring that the dual coordinate $y^\ast$ lies in the effective domain of the divergence $d_\phi$, and~\eqref{eq:exp_bregman_link} is a slight abuse of notation since $y^\ast$ may be $\pm\infty$, but nevertheless the divergence $d_\phi$ itself remains well defined on an appropriate extension of its domain.
See~\citet[Example 8]{banerjee2005clustering} for an example.
The special and uniquely symmetric case of squared Euclidean distance is obtained when $\phi(y) = \Vert y \Vert_2^2$.

\section{Adversarially robust generalized linear models}
\label{sec:adversarially-robust-posterior}

In this section, we derive the adversarially robust posterior $q_\delta(\theta)$ when the likelihood is respectively a Gaussian likelihood, and more generally an exponential family likelihood.
The result in~\Cref{lm:adv_loss_closed_form} may be of independent interest for studying adversarially robust models even in the setting of point-estimation.
It allows, for example, an adversarially robust extension of  logistic regression (binary-valued data), Poisson regression (count-valued data), and exponential or gamma regression (positive-valued data).
More generally, any generalized linear model~\citep{mccullagh1989generalized} with canonical link function may be adversarialized. 

\paragraph{Adversarial negative log likelihood} 
We consider adversarial losses $ {\ell}_\delta(\theta, \mathcal{D})$ of the form
\begin{align*}
     {\ell}_\delta\big(\theta, (x,y) \big) &=  \max_{\|  \widetilde{x} - x \|_2 \leq \delta} - \log p\big(y \mid  f_\theta(\widetilde{x}) \big) \numberthis \label{eq:nll_exponentialfam}\\
     &=  \max_{\|  \widetilde{x} - x \|_2 \leq \delta} d_\psi\big(f_\theta(\widetilde{x}), y^\ast \big) + C(y), \numberthis \label{eq:adv_loss}
\end{align*} 
where $\delta$ controls the allowable perturbation in the features. 
Considering a linear predictor and Gaussian likelihood (squared error Bregman divergence) allows us to derive the robust loss in closed-form. 
\begin{lemma}[Robust loss in closed-form for Gaussian likelihood] \label{lm:adv_loss_closed_form_gaussian}
Under a linear predictor $f_\theta(x)=\theta^\top x$ and in the case where the exponential family is a Gaussian family,
$$ \ell_\delta(\theta, \mathcal{D}) = \frac{n}{2}\log \lp 2\pi \sigma^2 \rp + \frac{1}{2\sigma^2} \Big\| |Y - X\theta| + \delta \|\theta\| 1_n \Big\|^2,$$
and the adversarial perturbation of the sample $x$ is $\widetilde{x} = \delta \sign(\theta^\top x - y) \frac{\theta}{\Vert \theta \Vert_2} + x = \delta \sign(\theta^\top \widetilde{x} - y) \frac{\theta}{\Vert \theta \Vert_2} + x$.
\end{lemma}
More generally, a linear predictor with any exponential family likelihood (Bregman divergence) allows us to derive the robust loss in closed-form.

\begin{lemma}[Robust loss in closed-form for exponential family likelihood]
\label{lm:robust-expfam-loss}
Under a linear predictor $f_\theta(x)=\theta^\top x$ and an exponential family likelihood, the robust loss is
    \begin{align*}
        &\phantom{{}={}} \ell_{\delta}\
\big(\theta, (x, y) \big) \\
&= \max_{s \in \{-1, 1\}}\psi(s \delta \Vert \theta \Vert_2 + \theta^\top x) - \psi(\theta^\top x) - y s \delta \Vert \theta \Vert_2 \\
&\phantom{{}=\max_{s \in \{-1, 1\}}}+ d_\psi(\theta^\top x, y^\ast) + C(y),
    \end{align*}
and the adversarial perturbation of the sample $x$ is $\widetilde{x} = \delta \sign \big(\nabla(\psi(\theta^\top \widetilde{x}) - y \big) \frac{\theta}{\Vert \theta \Vert_2} + x$.
\label{lm:adv_loss_closed_form}
\end{lemma}
Note that in the case of a general exponential family likelihood, a trivial maximization problem over $s \in \{-1, 1\}$ must be solved.
All other terms in~\Cref{lm:adv_loss_closed_form} are available in closed-form. 
In practice, this optimization problem can be solved by simply evaluating the objective for $s=1$ and $s=-1$, and picking the result with the highest value.
\Cref{lm:adv_loss_closed_form,lm:adv_loss_closed_form_gaussian} are proved in \Cref{app:adv_loss}. 

We are now ready to define our robust posterior for Bayesian generalized linear models.
\begin{corollary}[Robust posterior]    \label{cor:robust_posterior}
    The Gibbs posterior~\eqref{eq:gibbs_posterior} obtained by setting the loss $\mathcal{L}$ to be an adversarially perturbed exponential family NLL~\eqref{eq:nll_exponentialfam} (or equivalently, an adversarially perturbed Bregman divergence~\eqref{eq:adv_loss}) under a linear model $f_\theta(x) = \theta^\top x$ is given by
    \begin{align*}
        q_\delta(\theta) = \frac{\exp\Big( -\sum_{i=1}^N\ell_\delta\big( \theta, (x_i, y_i)\big) \Big) \pi(\theta) }{\int \exp\Big( - \sum_{i=1}^N\ell_\delta\big( \theta', (x_i, y_i)\big) \Big) \pi(\theta') d\theta'},
    \end{align*}
    where $\ell_\delta\big( \theta, (x_i, y_i)\big) $ is as in~\Cref{lm:adv_loss_closed_form}, or in the special case of a Gaussian (or squared loss),~\Cref{lm:adv_loss_closed_form_gaussian}.
\end{corollary}
We note that this notion of a robust posterior is not the only choice, however this choice does lead to tractable losses derived from adversarial likelihoods, and also allows us to derive generalization guarantees in \Cref{sec:pac_bayes_certs}.
See \Cref{app:adv_loss} for a discussion of other choices.
\section{Standard and Adversarial Generalization Certificates}
\label{sec:pac_bayes_certs}
In this section, we focus on Bayesian linear regression (i.e. a robustified squared error loss or Gaussian NLL) for the robust posterior in~\Cref{cor:robust_posterior}.
We consider labels generated using a true parameter $\theta^*$,  $y_i=x_i^\top \theta^* + \epsilon_i$ where $\mathbb{E}[x_i] = 0, \, \mathbb{E}[\|x_i\|^2] = \sigma_x^2$ and $\epsilon_i \sim \mathcal{N}(0,\sigma^2)$. 

\paragraph{PAC-Bayesian generalization certificates} 
Unlike traditional generalization bounds based on uniform convergence such as VC-dimension \citep{vapnik1971uniform}, Rademacher complexity \citep{shalev2014understanding}, and information-theory \citep{zhang2006information}, PAC-Bayes \citep{mcallester1998some} focuses on Bayesian predictors 
rather than a single deterministic hypothesis class. This perspective allows PAC-Bayes to provide \emph{data-dependent} generalization guarantees, 
that are computed on training samples without relying on the test data. 
As such, all certificates computed in this section depend on the data $X$, $Y$ in a non-obvious way.
While other approaches based on uniform convergence or information theory result in a worst-case guarantee, PAC-Bayesian offers fine-grained analysis by taking advantage of informed prior choice leading to a tighter certificate. 

\paragraph{Data and constants in bounds} In addition to the data $X, Y$, each of the bounds in this section also rely on constants such as the training and testing perturbation allowances $\delta$ and $\deltat$.
We intuitively describe the role of these data and constant terms following the presentation of each of the bounds, even though the main purpose of the bounds is as computable data-dependent certificates.

\paragraph{Standard and adversarial generalization}
We derive certificates for both standard and adversarial generalization (columns in \Cref{tab:overview}). In order to formalize this, let us define the expected and empirical errors for standard loss $\ell$ as $R(\theta)=\expectation{(x,y) \sim \mathcal{P}}{\ell\big(\theta, (x,y)\big)}$ and $r(\theta)=\frac{1}{n} \mathcal{L}(\theta, \mathcal{D})$, respectively. Similarly, the expected and empirical adversarial errors under perturbation $\delta$ are defined as $ {R}_\delta(\theta)=\expectation{(x,y) \sim \mathcal{P}}{ {\ell}_\delta\big(\theta, (x,y)\big)}$ and $ {r}_\delta(\theta)=\frac{1}{n}  \mathcal{L}_\delta(\theta, \mathcal{D})$, respectively. 
In the context of Bayesian inference, the standard generalization risk certificate quantifies the expected loss of the posterior $\rho(\theta)$ on unperturbed test data $x$: $\expectation{\theta \sim \rho}{R(\theta)}$. 
Similarly, the adversarial generalization certificate quantifies the performance
of the test data under adversarial perturbation $\deltat$: 
$\expectation{\theta \sim \rho}{ {R}_{\deltat}(\theta)}$. 
We derive the certificates by bounding the respective quantity for both the standard Bayes posterior $q(\theta)$ and adversarially robust posterior $ {q}_\delta(\theta)$.
Note that we allow the perturbation $\delta$ used for inference (i.e. calculation of $ {q}_\delta(\theta)$) to be different to the perturbation $\deltat$ at test-time.

\subsection{Cumulant generating function}

To derive the standard and adversarial generalization certificates, we leverage the PAC-Bayesian theorem for any loss with bounded cumulant generating function (CGF) in \citet{banerjee2021information}. 
We state the result in \Cref{th:pac_bayes_bounded_cgf}, which requires a bounded CGF.
We then show that the CGFs of the standard and adversarial losses corresponding with Gaussian NLLs are bounded in \Cref{lm:cgf_bounds_std_adv_loss}. 
\begin{theorem}[Theorem 6 in \citet{banerjee2021information}] \label{th:pac_bayes_bounded_cgf}
Consider data $\mathcal{D}$ and any loss $\ell \big(\theta, (x,y)\big)$ with its corresponding expected and empirical generalization errors $R(\theta)=\expectation{(x,y) \sim \mathcal{P}}{\ell \big(\theta, (x,y)\big)}$ and $r(\theta)=\frac{1}{n} \mathcal{L}(\theta, \mathcal{D})$, respectively.  
Let the CGF of the loss $\psi(t) = \log \expectation{}{\exp \lp t \lp \expectation{}{\ell } - \ell  \rp \rp}$  be bounded, where for some constant $c>0$, $t \in (0,1/c)$. 
Then, we have, with probability at least $1-\beta$, for all densities 
$\rho(\theta)$,
    $$\expectation{\theta \sim \rho}{R(\theta)}
\leq 
\expectation{\theta \sim \rho}{r(\theta)} 
+ \frac{1}{t} \left[ \frac{\mathrm{KL}({\rho} \| \pi) + \log \frac{1}{\beta}}{n} + \psi(t) \right].$$
\end{theorem}
Before stating the CGF bound for the losses, we first define a \textit{sub-gamma} random variable. A random variable with variance $s^2$ and scale parameter $c$ is said to be sub-gamma if its CGF $\psi$ satisfies the following upper bound: 
$$\psi(t) \leq \frac{s^2t^2}{2(1-ct)} \text{ for all } 0 < t < 1/c.$$ 
We state the CGF bounds for both standard and adversarial losses in \Cref{lm:cgf_bounds_std_adv_loss} and provide the proof in \Cref{app:cgf_std_adv_loss}.

\begin{lemma}
[CGF bounds for standard and adversarial losses] \label{lm:cgf_bounds_std_adv_loss}
The standard and adversarial losses are both sub-gamma with the following variance $s^2$ and scale factor $c$. 
In the case of standard loss, 
\begin{align}
    c&=\frac{\sigma^2_p \sigma_x^2}{\sigma^2}, \quad 
    s^2 = \frac{1}{t} \lp cd - ct +1 + \frac{\sigma_x^2 \|\theta^*\|^2}{\sigma^2}\rp.
    \label{eq:cgf_std_loss}
\end{align}
For adversarial loss with $\deltat$ perturbation, 
\begin{align}
c&=\frac{2\sigma^2_p\lp \sigma_x^2+\deltat^2 \rp}{\sigma^2}, \nonumber \\
s^2 &= \frac{2}{t} \lp cd - ct + 1 + \frac{\sigma_x^2 \|\theta^*\|^2}{\sigma^2} \rp.
\label{eq:cgf_adv_loss}
\end{align}
\end{lemma}

Using the sub-gamma property of the losses and applying their CGF bounds in \Cref{th:pac_bayes_bounded_cgf}, we derive the standard and adversarial generalizations of Bayes posterior $q(\theta)$ in \Cref{ss:bayes_cert}, 
and robust posterior $q_\delta(\theta)$ in \Cref{ss:robust_cert}, and present the proofs in \Cref{app:pac_bounds_proof_robust_posterior}. 
Each of the bounds depends on parameters $c$ and $s^2$, and intuitively larger values of either lead to worse bounds, as the losses are subject to higher variability.

\subsection{Generalization certificates for Bayes posterior}
\label{ss:bayes_cert}
Using the sub-gamma property of the standard loss, the certificate for the standard generalization of the Bayes posterior $q(\theta)$ is derived in \citet{germain2016pac} by setting the free variable $t$ in CGF to $1$. We restate this result in \Cref{thm:std_post_std_loss}, expressing it explicitly in terms of the data. This contrasts with the formulation in \citet[Corollary 5]{germain2016pac}, where the bound is expressed in terms of the posterior (which in turn depends on the data). 

\begin{theorem}[Standard generalization of Bayes posterior, adapted from \citet{germain2016pac}] \label{thm:std_post_std_loss}
Consider $c$ and $s$ as defined in \Cref{eq:cgf_std_loss} with $t=1$, $\sigma_p^2 < \frac{\sigma^2}{\sigma_x^2}$, $W_d= I_d + \frac{\sigma_p^2}{\sigma^2} X^\top X$ and $W_n= I_n + \frac{\sigma_p^2}{\sigma^2} XX^\top$. Then, with probability at least $1-\beta$, we have the following certificate for standard generalization of the Bayes posterior $q(\theta)$: 
\begin{align}
\expectation{\theta \sim q}{R(\theta)} &\leq \frac{1}{n} \log \sqrt{\det \lp W_d \rp} + \frac{1}{2n\sigma^2} Y^\top W_n^{-1} Y  \nonumber \\
&\qquad + \frac{1}{n} \log \frac{1}{\beta}  +\frac{s^2}{2(1-c)}.
\label{eq:pac_bayes_std_post_std_loss}
\end{align}
\end{theorem}
Increasing sub-gamma variability parameters $s^2$ and $c$ increase the bound~\eqref{eq:pac_bayes_std_post_std_loss}, as expected. 
Informally, the term depending on $\log \det W_d$ is a sum of $d$ $\log$ eigenvalues of $W_d$, and if $X^\top X$ is low-rank, most of these $\log$ eigenvalues are close to $0$. 
Hence the first term decreases like $1/n$.
The other data dependent term is essentially the product of $Y^\top$ and the average training error of ridge regression, which should be small if the dataset is large and the model is well-specified.

Next, we derive the adversarial generalization certificate for Bayes posterior similar to standard generalization.

\begin{theorem}[Adversarial generalization of Bayes posterior]\label{thm:std_post_adv_loss}
Consider $c$ and $s$ as defined in \Cref{eq:cgf_adv_loss} with $t=1$, $\sigma_p^2 < \frac{\sigma^2}{2 \lp \sigma_x^2 + \deltat^2\rp}$, $W_d= I_d + \frac{\sigma_p^2}{\sigma^2} X^\top X$ and $W_n= I_n + \frac{\sigma_p^2}{\sigma^2} XX^\top$. Then, with probability at least $1-\beta$, we have the following certificate for adversarial generalization of the Bayes posterior ${q}(\theta)$: 
\begin{align}
&\expectation{\theta\sim q}{  {R}_{\deltat}(\theta)} \leq \frac{2}{n} \log \sqrt{\det \lp W_d \rp} + \frac{1}{n\sigma^2} Y^\top W_n^{-1} Y \nonumber \\
&\qquad \qquad + \frac{1}{n} \log \frac{1}{\beta}  +\frac{s^2}{2(1-c)} + \frac{d\deltat^2 \sigma_p^2}{\sigma^2-2n\deltat^2 \sigma_p^2}. 
\label{eq:pac_bayes_std_post_adv_loss}
\end{align}
\end{theorem}

While \Cref{thm:std_post_std_loss,thm:std_post_adv_loss} are upper bounds and are incomparable, we note that the main difference in \Cref{thm:std_post_adv_loss} is the additional constant term dependent on the perturbation radius $\deltat$ and the data-dependent terms are scaled by $2$. 
The additional constant term captures the effect of testing the model adversarially, increasing the bound. 
This term behaves linearly in $\deltat$ for small $\deltat$.

\subsection{Generalization certificates for robust posterior}
\label{ss:robust_cert}

First we derive the standard generalization of robust posterior. While this setting may not be of practical interest, we provide the result for completeness.

\begin{theorem}[Standard generalization of robust posterior] \label{thm:adv_post_std_loss}
Consider $c$ and $s$ as defined in \Cref{eq:cgf_std_loss} with $t=1$, $\sigma_p^2 < \frac{\sigma^2}{\sigma_x^2}$, 
$k_\delta = \frac{2n\delta^2\sigma_p^2}{\sigma^2} + 1$, 
$U_d= k_\delta I_d + \frac{2\sigma_p^2}{\sigma^2} X^\top X$,
$U_n= k_\delta I_n + \frac{2\sigma_p^2}{\sigma^2} XX^\top$, 
$V_d= k_\delta I_d + \frac{\sigma_p^2}{\sigma^2} X^\top X$, and $V_n= k_\delta I_n + \frac{\sigma_p^2}{\sigma^2} X X^\top$. Then, with probability at least $1-\beta$, we have the following certificate for standard generalization of the robust posterior ${q}_\delta(\theta)$: 
\begin{align}
\expectation{  \theta \sim {q}_\delta}{R(\theta)}  &\leq \frac{2}{n} \log \sqrt{\det(U_d)} + \frac{2}{n k_\delta \sigma^2}Y^\top U_n^{-1} Y \nonumber \\
&\quad - \frac{1}{n} \log \sqrt{\det \lp V_d \rp} -  \frac{k_\delta}{n \sigma^2} Y^\top V_n^{-1} Y \nonumber \\
&\quad + \frac{1}{n} \log \frac{1}{\beta}  +\frac{s^2}{2(1-c)}. 
\label{eq:pac_bayes_adv_post_std_loss}
\end{align}
\end{theorem}
The terms involving $\det U_d$ and $\det V_d$ are sums of $d$ $\log$ eigenvalues divided by $n$, so they scale like $1 / n$. 
As in the previous bounds, the remaining data dependent bounds resemble the product of $Y^\top$ and the average error of ridge regression with an effective regularization parameter.

For the robust posterior, we consider the cases $\deltat=\delta$ and $\deltat\neq\delta$ separately. We derive a tighter bound for the special case when the allowed adversarial perturbation at train and test-time are the same, i.e., $\deltat = \delta$. 
\begin{theorem}[Adversarial generalization of robust posterior with $\deltat=\delta$] \label{thm:adv_post_adv_loss}
Consider $c$ and $s$ as defined in \Cref{eq:cgf_adv_loss} with $t=1$, $\sigma_p^2 < \frac{\sigma^2}{2 \lp \sigma_x^2 + \deltat^2\rp}$,
$k_\delta = \frac{2n\delta^2\sigma_p^2}{\sigma^2} + 1$, 
$U_d= k_\delta I_d + \frac{2\sigma_p^2}{\sigma^2} X^\top X$, and $U_n= k_\delta I_n + \frac{2\sigma_p^2}{\sigma^2} XX^\top$. 
Then, with probability at least $1-\beta$, we have the following certificate for adversarial generalization of the robust posterior $  {q}_\delta(\theta)$: 
\begin{align}
\expectation{  \theta \sim {q}_\delta}{  {R}_\delta(\theta)}  &\leq \frac{1}{n} \log \sqrt{\det \lp U_d \rp} + \frac{1}{n k_\delta \sigma^2} Y^\top U_n^{-1} Y  \nonumber \\
&\qquad + \frac{1}{n} \log \frac{1}{\beta}  +\frac{s^2}{2(1-c)}. \label{eq:pac_bayes_adv_post_adv_loss}
\end{align}
\end{theorem}
Compared with \Cref{thm:adv_post_std_loss}, \Cref{thm:adv_post_adv_loss} only includes $1$ times the $U_d$ and $U_n$ dependent terms, instead of $2$ times the $U_d$ and $U_n$ dependent terms minus the $V_d$ and $V_d$ dependent terms. 
Empirically, in \Cref{sec:exp}, we find that this leads to a favorable bound.

Finally, using a different analysis we derive the adversarial generalization focusing on a general setting where the adversarial  perturbation radius at test-time $\deltat$ is not the same as the radius used to learn the posterior $\delta$.
\begin{theorem}[Adversarial generalization of robust posterior] \label{thm:adv_post_adv_loss_gen}
Consider $c$ and $s$ as defined in \Cref{eq:cgf_adv_loss} with $t=1$, $\sigma_p^2 < \frac{1}{4 \lp \sigma_x^2 + \deltat^2\rp}$,
$k_\delta = \frac{2n\delta^2\sigma_p^2}{\sigma^2} + 1$, 
$U_d= k_\delta I_d + \frac{2\sigma_p^2}{\sigma^2} X^\top X$, and $U_n= k_\delta I_n + \frac{2\sigma_p^2}{\sigma^2} XX^\top$. Then, with probability at least $1-\beta$, we have the certificate for adversarial generalization of the robust posterior $  {q}_\delta(\theta)$: 
\begin{align}
&\expectation{  \theta \sim {q}_\delta}{  {R}_{\deltat}(\theta)} \leq \frac{2}{n} \log \sqrt{\det \lp U_d \rp} + \frac{2}{n k_\delta \sigma^2} Y^\top U_n^{-1} Y  \nonumber \\
&\,\, + \frac{1}{n} \log \frac{1}{\beta}  +\frac{s^2}{2(1-c)} + \frac{\lp \deltat^2 -\delta^2 \rp \sigma_p^2 d}{\sigma^2-2n\lp \deltat^2 -\delta^2 \rp \sigma_p^2}.
\label{eq:pac_bayes_adv_post_adv_loss_gen}
\end{align}
\end{theorem}


\section{Experimental Results}
\label{sec:exp}

\begin{table*}[t]
    \centering
    \begin{tabular}{lccccc}
    \toprule
    \multirow{2}{*}{Dataset}  
       & \multicolumn{2}{c}{Standard generalization (NLL) $\ell$} & \phantom{}& \multicolumn{2}{c}{Adversarial generalization (adv-NLL) $\ell_{\deltat}$} \\
       & Bayes posterior $q$  & Robust posterior  $q_{\delta}$ & \phantom{}& Bayes posterior $q$ & Robust posterior $q_{\delta}$  \\
       \midrule
       Abalone & \textbf{1.1586} \tiny{$\pm$ 0.013} & 1.1729 \tiny{$\pm$ 0.015} && 1.2539 \tiny{$\pm$ 0.012} & \textbf{1.2178} \tiny{$\pm$ 0.015} \\
       Air Foil & \textbf{1.1656} \tiny{$\pm$ 0.008} & 1.1690 \tiny{$\pm$ 0.008} && 1.2194 \tiny{$\pm$ 0.008} & \textbf{1.2175} \tiny{$\pm$ 0.008} \\
       Air Quality & \textbf{0.9665} \tiny{$\pm$ 0.002} & 0.9670 \tiny{$\pm$ 0.002} && 0.9826 \tiny{$\pm$ 0.003} & \textbf{0.9792} \tiny{$\pm$ 0.002} \\
       Auto MPG & 1.0231 \tiny{$\pm$ 0.006} & \textbf{1.0228} \tiny{$\pm$ 0.007} && 1.0552 \tiny{$\pm$ 0.006} & \textbf{1.0469} \tiny{$\pm$ 0.008} \\
       California Housing & \textbf{1.1193} \tiny{$\pm$ 0.003} & 1.1267 \tiny{$\pm$ 0.005} && 1.1910 \tiny{$\pm$ 0.003} & \textbf{1.1769} \tiny{$\pm$ 0.005} \\
       Energy Efficiency & \textbf{0.9709} \tiny{$\pm$ 0.006} & 0.9731 \tiny{$\pm$ 0.007} && 0.9996 \tiny{$\pm$ 0.007} & \textbf{0.9945} \tiny{$\pm$ 0.008} \\
       Wine Quality & 1.2339 \tiny{$\pm$ 0.007} & \textbf{1.2322} \tiny{$\pm$ 0.006} && 1.2696 \tiny{$\pm$ 0.008} & \textbf{1.2627} \tiny{$\pm$ 0.006} \\
       \bottomrule
    \end{tabular}
    \caption{
    Test NLL and adversarial NLL of Bayes and robust posteriors on real datasets. The prior variance is set to $\sigma_p^2 = \frac{1}{d}$. The robust posterior is trained with $\delta=0.1$, and adversarial generalization is evaluated using the same training-time perturbation ($\deltat=0.1$). The adversarial generalization results demonstrate that the robust posterior $q_\delta$ is consistently more robust than the Bayes posterior $q$. For both standard and adversarial generalization, the best-performing model is bold.
       \label{tab:real_data} }
       \vspace{-0.3cm}
\end{table*}

In this section, we present $(i)$ adversarial robustness of Bayes and robust posteriors on real datasets; $(ii)$ validation of the derived generalization certificates for the posteriors, and compare it to the prior work of \citep{germain2016pac}.
Code to reproduce all experiments is provided~\footnote{\url{https://figshare.com/s/dc9034bb2e323a87a7a4}}, and implementation and hardware details are given in \Cref{app:exp}.

\paragraph{Datasets and hyperparameters} 
We consider the following regression datasets with $70-30$ train-test split: \emph{Abalone} \citep{abalone}, \emph{Air Foil} \citep{airfoil}, \emph{Air Quality} \citep{air_quality}, \emph{Auto MPG} \citep{auto_mpg}, \emph{California Housing} \citep{pedregosa2011cali}, \emph{Energy Efficiency} \citep{energy_efficiency}, \emph{Wine Quality} \citep{wine_quality}. The datasets are standardized to zero mean and unit variance. 
We provide results for prior variance $\sigma_p^2 = \{\frac{1}{100}, \frac{1}{9}, \frac{1}{d}\}$ where $d$ is the data feature dimension. 
For certificate validation, we use a synthetic dataset where the data features $x \sim \mathcal{N}(0, \sigma_x^2 I_d)$ and $y = \theta^* x^\top + \epsilon$  with $d=5$ and $\epsilon \sim \mathcal{N}(0, \sigma^2)$ with $\sigma^2 = \frac{1}{9}$ and $\| \theta^* \|^2 = 0.5$. We fix prior variance $\sigma_p^2 = 0.01$ and consider a range of training samples from $10$ to $10^4$ and $10^4$ test samples. 
All results are averaged over 5 seeds and reported with standard deviation.
We employ Hamiltonian Monte Carlo (HMC) to efficiently sample from the posterior distribution. Specifically, we utilize the No-U-Turn Sampler (NUTS) \citep{hoffman2014no}, an adaptive variant of HMC that automatically tunes step sizes and trajectory lengths for improved sampling efficiency. 

\begin{figure*}[t]
    \centering\vspace{-0.1cm}
    \includegraphics[width=0.24\linewidth]{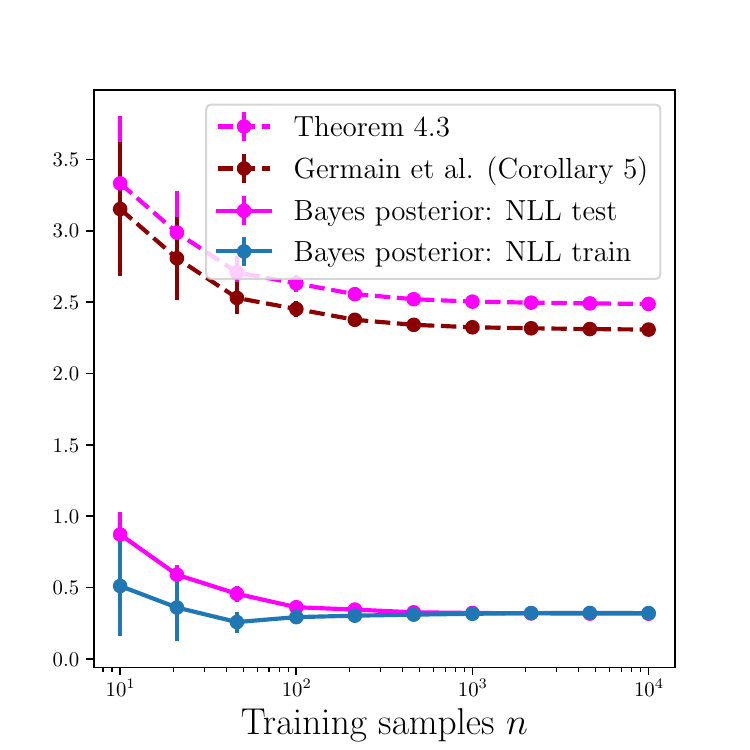}
    \includegraphics[width=0.24\linewidth]{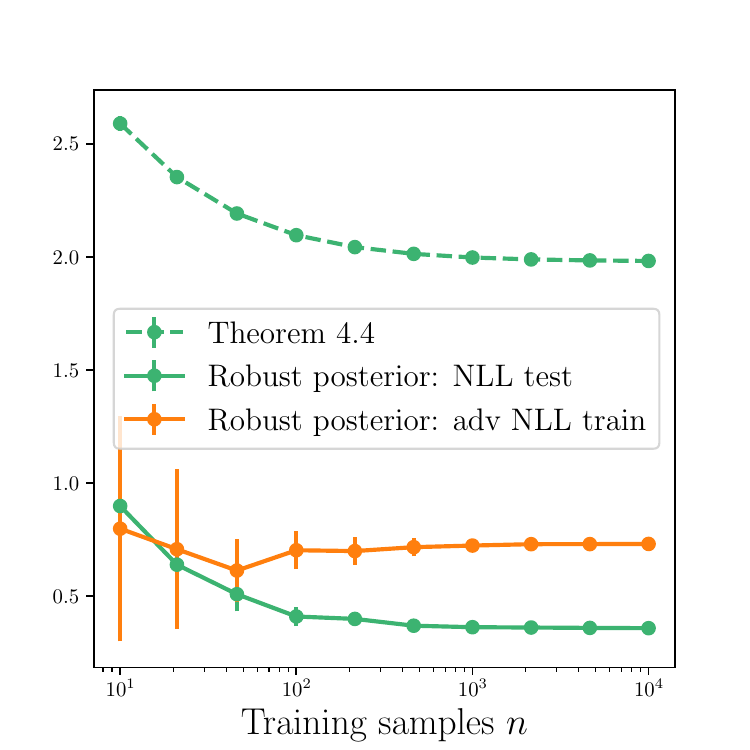}
    \includegraphics[width=0.24\linewidth]{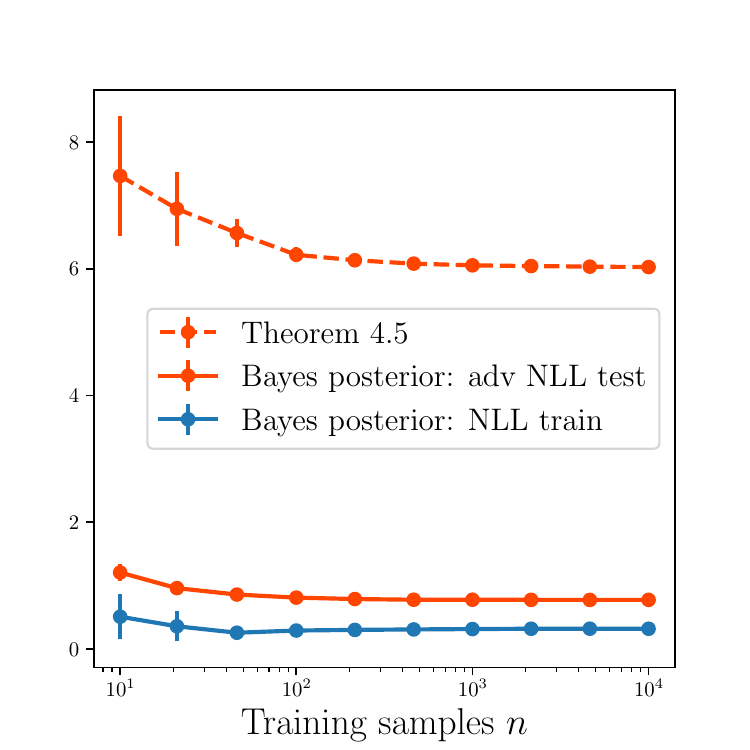}
    \includegraphics[width=0.24\linewidth]{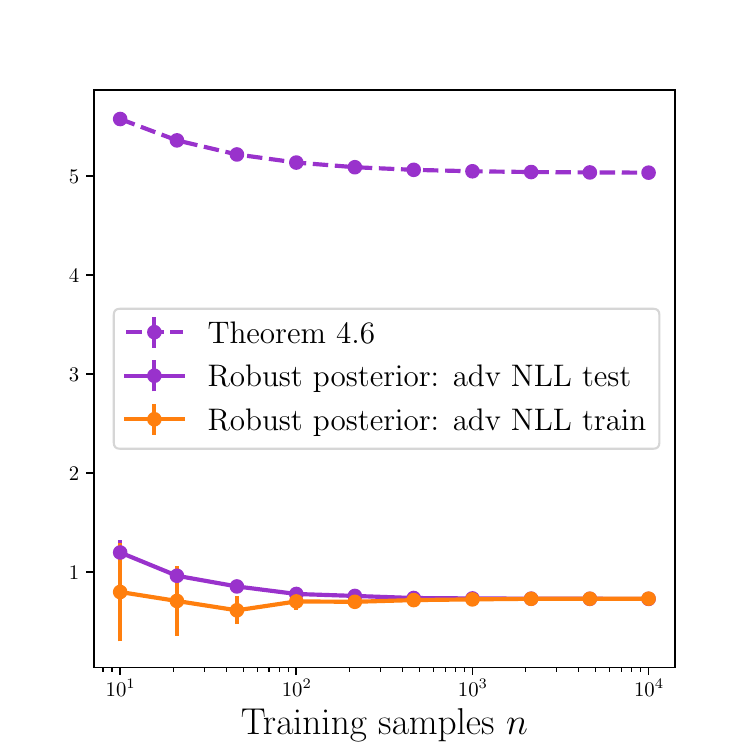}
    \caption{Validation of the derived generalization certificates \Cref{thm:std_post_std_loss,thm:std_post_adv_loss,thm:adv_post_std_loss,thm:adv_post_adv_loss}. (left to right) Standard generalization of Bayes posterior with comparison to prior work \citet{germain2016pac}, standard generalization of the robust posterior, adversarial generalization of Bayes posterior, and adversarial generalization of the robust posterior. 
    \label{fig:pac_bounds}}
    \vspace{-0.3cm}
\end{figure*}

\subsection{Results on real data}

We present the standard and adversarial generalization results for Bayes and robust posteriors evaluated on real data  in \Cref{tab:real_data}, using a prior variance of $\sigma_p^2 = \frac{1}{d}$. Additional results for $\sigma_p^2 = \{\frac{1}{100}, \frac{1}{9}\}$ are in \Cref{app:exp_real}. 
Our findings clearly demonstrate that \emph{the robust posterior $q_\delta$ consistently outperforms the standard Bayes posterior $q$ in terms of adversarial robustness}. Moreover, for certain choices of prior, the robust posterior also achieves superior standard generalization (see \Cref{tab:real_data,tab:real_data_p9,tab:real_data_p100}). 
This suggests that adversarial robustness in Bayesian models can potentially be enhanced through probabilistic inference and may not always be subject to the severe robustness-accuracy trade-off commonly observed in non-Bayesian models \citep{tsipras2019robustness}.

\subsection{Validation of certificates} 
We validate our derived generalization certificates in \Cref{fig:pac_bounds} by plotting the PAC-Bayesian bounds as a function of the number of training samples. Since these bounds provide rigorous upper estimates on the generalization error, they are not directly comparable to each other but rather serve as theoretical guarantees. While the bounds may appear conservative, it is important to note that these are the first rigorous PAC-Bayesian bounds for adversarial robustness.
Additionally, we compare our standard generalization bound for the Bayes posterior with the prior work \citep{germain2016pac} in \Cref{fig:pac_bounds} (left). Although both approaches leverage PAC-Bayesian principles, the bound from \citep{germain2016pac} is numerically lower than ours because they approximate the expected training loss in  \Cref{th:pac_bayes_bounded_cgf} using the empirical loss, while we compute the actual expected training loss. 

\section{Discussion and Related Work}
\label{sec:related_works}


\paragraph{Generalization bounds for probabilistic models} 
Estimating the average generalization performance as the availability of training data increases, commonly referred to as learning curves, has been extensively studied for probabilistic models such as Gaussian processes (GPs) and Bayesian linear models. 
In GP regression, significant strides have been made in understanding generalization performance over the past two decades. For instance, \citet{sollich2002learning, sollich1998learning} estimated learning curves by bounding the prediction variance, while \citet{opper1998general} analyzed learning curves through bounds on prediction error. Further developments include \citet{sollich2001gaussian}, who investigated learning curves under mismatched models, 
and \citet{jin2022learning}, who provided a more realistic analysis by assuming the eigenspectrum of the prior and the eigenexpansion coefficients of the target function follow a power-law distribution.
Notably, \citet{williams2000upper} derived non-trivial upper and lower bounds for GPs, offering key insights into their generalization capabilities. More recently, \citet{savvides2024error} bounded both the variance of the predictions and the bias.
While learning curves provide valuable average-case insights, they do not offer certificates.

\paragraph{Generalization certificates and their practical significance}
In the context of GPs, PAC-Bayesian bounds are derived for GP classification  \citet{seeger2002pac} and GP regression \citep{suzuki2012pac}. 
Beyond GPs, PAC-Bayesian theory has been instrumental in explaining the generalization capabilities of neural networks where the uniform convergence type of analyses fail \citep{dziugaite2017computing,lotfi2022pac}.
Practically, these bounds are useful in hyperparameter optimization \citep{cherian2020efficient} and improving model training \citep{reeb2018learning,wang2023improving} by directly minimizing the bound.

\paragraph{Robustness certificates for GPs} 
While guaranteeing robustness against adversarial perturbations remain a relatively underexplored area, it is important to note that even heuristic methods aimed at improving the robustness of probabilistic models are less developed \citep{hernandez2011robust,bradshaw2017adversarial,grosse2018limitations}. Moreover, there exists no notion of adversarially robust posteriors or GPs. In the context of GP classification, \citet{blaas2020adversarial} provided robustness guarantees for the standard Bayes posterior by computing upper and lower bounds for the maximum and minimum of GP classification probabilities under adversarial perturbations. 
Similar analysis of robustness certification for standard GP regression has been investigated in works such as \citet{patane2022adversarial} and \citet{cardelli2019robustness}. 

\paragraph{Other robust posteriors} 
In addition to alternative robust posteriors under the optimization-centric view of Bayes rule (as discussed in Appendix~\ref{app:adv_loss}) existing and distinct notions of robust posterior are also available.
Focusing on the categorical distribution (a special exponential families), \citet{wicker2021bayesian} define a robust likelihood by marginalizing out perturbed softmax probability distributions with respect to a distribution on the perturbation allowance. 
This is then used as a standard likelihood inside Bayesian inference to building Bayesian neural networks.

\paragraph{Adversarially robust optimization} 
Although adversarially robust optimization might appear conceptually similar to our proposed adversarially robust posterior formalism, it addresses a fundamentally different problem. In Bayesian optimization, the goal is to select $x_t$ such that it yields a high value even under adversarial perturbations, i.e., maximizing $f(\tilde{x}_t)$, where $f$ is an unknown function \citep{bogunovic2018adversarially, kirschner2020distributionally}. The fundamental distinction lies in the fact that $f$ is not explicitly known in adversarially robust optimization. 
Moreover, the objective of sequentially choosing $x$ to learn the unknown $f$ is different from learning a posterior from a given data that is adversarially robust.
\section{Conclusion}
\label{sec:conclusion}

We consider the problem of adversarially robust probabilistic inference.
Using the generalized Bayesian inference framework, we propose adversarially robust posteriors.
We show that for exponential family models, closed-form  adversarial NLLs 
result in posteriors that are robust to adversarial perturbations.
We derive PAC-Bayes generalization bounds for the four cases as summarized in \Cref{tab:overview}.
The $2\times 2$ table corresponds to combinations of the following settings:
$(i)$ standard NLL $\ell$, and adversarial NLL $\ell_{\deltat}$;
$(ii)$ the classical Bayes posterior $q$, and the robust posterior $q_\delta$.
Our experiments validate that the derived PAC-Bayes bounds capture the empirical behavior, and demonstrate that the robust posterior $q_\delta$ consistently improves adversarial robustness.

Our work primarily focuses on Bayesian linear regression, with the exception of \Cref{lm:robust-expfam-loss} which may be of independent interest. 
This result opens the possibility of extending our analysis to other generalized linear models. 
We hope that our notion of adversarially robust posterior
will lead to further results in other machine learning problems, and provide useful analysis
for practical adversarial learning tasks in the real world.

\clearpage

\begin{acknowledgements} 
This work was partially done when MS visited Data61, CSIRO, and 
the authors would like to thank Peter Caley for facilitating the visit to Australia. Additionally, this research has been supported by the TUM Georg Nemetschek Institute Artificial Intelligence for the Built World. 
\end{acknowledgements}

\bibliography{citations}

\begin{thebibliography}{60}
\providecommand{\natexlab}[1]{#1}
\providecommand{\url}[1]{\texttt{#1}}
\expandafter\ifx\csname urlstyle\endcsname\relax
  \providecommand{\doi}[1]{doi: #1}\else
  \providecommand{\doi}{doi: \begingroup \urlstyle{rm}\Url}\fi

\bibitem[Alquier et~al.(2016)Alquier, Ridgway, and Chopin]{alquier2016properties}
Pierre Alquier, James Ridgway, and Nicolas Chopin.
\newblock On the properties of variational approximations of gibbs posteriors.
\newblock \emph{Journal of Machine Learning Research}, 17\penalty0 (236):\penalty0 1--41, 2016.

\bibitem[Banerjee et~al.(2005)Banerjee, Merugu, Dhillon, and Ghosh]{banerjee2005clustering}
Arindam Banerjee, Srujana Merugu, Inderjit~S Dhillon, and Joydeep Ghosh.
\newblock Clustering with bregman divergences.
\newblock \emph{Journal of machine learning research}, 6\penalty0 (10), 2005.

\bibitem[Banerjee and Mont{\'{u}}far(2021)]{banerjee2021information}
Pradeep~Kr. Banerjee and Guido Mont{\'{u}}far.
\newblock Information complexity and generalization bounds.
\newblock In \emph{{IEEE} International Symposium on Information Theory, {ISIT}}, 2021.

\bibitem[Bingham et~al.(2019)Bingham, Chen, Jankowiak, Obermeyer, Pradhan, Karaletsos, Singh, Szerlip, Horsfall, and Goodman]{bingham2019pyro}
Eli Bingham, Jonathan~P. Chen, Martin Jankowiak, Fritz Obermeyer, Neeraj Pradhan, Theofanis Karaletsos, Rohit Singh, Paul~A. Szerlip, Paul Horsfall, and Noah~D. Goodman.
\newblock Pyro: Deep universal probabilistic programming.
\newblock \emph{Journal of Machine Learning Research}, 2019.

\bibitem[Bishop(2007)]{bishop2007pattern}
Christopher~M. Bishop.
\newblock \emph{Pattern recognition and machine learning, 5th Edition}.
\newblock Information science and statistics. Springer, 2007.
\newblock ISBN 9780387310732.

\bibitem[Blaas et~al.(2020)Blaas, Patane, Laurenti, Cardelli, Kwiatkowska, and Roberts]{blaas2020adversarial}
Arno Blaas, Andrea Patane, Luca Laurenti, Luca Cardelli, Marta Kwiatkowska, and Stephen Roberts.
\newblock Adversarial robustness guarantees for classification with gaussian processes.
\newblock In \emph{International Conference on Artificial Intelligence and Statistics}. PMLR, 2020.

\bibitem[Bogunovic et~al.(2018)Bogunovic, Scarlett, Jegelka, and Cevher]{bogunovic2018adversarially}
Ilija Bogunovic, Jonathan Scarlett, Stefanie Jegelka, and Volkan Cevher.
\newblock Adversarially robust optimization with gaussian processes.
\newblock \emph{Advances in neural information processing systems}, 2018.

\bibitem[Bradshaw et~al.(2017)Bradshaw, Matthews, and Ghahramani]{bradshaw2017adversarial}
John Bradshaw, Alexander G de~G Matthews, and Zoubin Ghahramani.
\newblock Adversarial examples, uncertainty, and transfer testing robustness in gaussian process hybrid deep networks.
\newblock \emph{arXiv preprint arXiv:1707.02476}, 2017.

\bibitem[Brooks et~al.(1989)Brooks, Pope, and Marcolini]{airfoil}
Thomas Brooks, D.~Pope, and Michael Marcolini.
\newblock {Airfoil Self-Noise}.
\newblock UCI Machine Learning Repository, 1989.

\bibitem[Cardelli et~al.(2019)Cardelli, Kwiatkowska, Laurenti, and Patane]{cardelli2019robustness}
Luca Cardelli, Marta Kwiatkowska, Luca Laurenti, and Andrea Patane.
\newblock Robustness guarantees for bayesian inference with gaussian processes.
\newblock In \emph{Proceedings of the AAAI conference on artificial intelligence}, 2019.

\bibitem[Catoni and Picard(2004)]{catoni2004statistical}
Olivier Catoni and Jean Picard.
\newblock Statistical learning theory and stochastic optimization ecole d'et{\'e} de probabilit{\'e}s de saint-flour xxxi-2001.
\newblock 2004.

\bibitem[Cherian et~al.(2020)Cherian, Taube, McGibbon, Angelikopoulos, Blanc, Snarski, Richman, Klepeis, and Shaw]{cherian2020efficient}
John~J. Cherian, Andrew~G. Taube, Robert~T. McGibbon, Panagiotis Angelikopoulos, Guy Blanc, Michael Snarski, Daniel~D. Richman, John~L. Klepeis, and David~E. Shaw.
\newblock Efficient hyperparameter optimization by way of {PAC-Bayes} bound minimization.
\newblock \emph{CoRR}, abs/2008.06431, 2020.
\newblock URL \url{https://arxiv.org/abs/2008.06431}.

\bibitem[Cortez et~al.(2009)Cortez, Cerdeira, Almeida, Matos, and Reis]{wine_quality}
Paulo Cortez, A.~Cerdeira, F.~Almeida, T.~Matos, and J.~Reis.
\newblock {Wine Quality}.
\newblock UCI Machine Learning Repository, 2009.

\bibitem[Csisz{\'a}r(1975)]{csiszar1975divergence}
Imre Csisz{\'a}r.
\newblock I-divergence geometry of probability distributions and minimization problems.
\newblock \emph{The annals of probability}, 1975.

\bibitem[Deisenroth et~al.(2020)Deisenroth, Faisal, and Ong]{deisenroth20mathml}
Marc~Peter Deisenroth, A~Aldo Faisal, and Cheng~Soon Ong.
\newblock \emph{Mathematics for Machine Learning}.
\newblock Cambridge University Press, 2020.

\bibitem[Donsker and Varadhan(1983)]{donsker1983asymptotic}
Monroe~D Donsker and SR~Srinivasa Varadhan.
\newblock Asymptotic evaluation of certain markov process expectations for large time. iv.
\newblock \emph{Communications on pure and applied mathematics}, 1983.

\bibitem[Dziugaite and Roy(2017)]{dziugaite2017computing}
Gintare~Karolina Dziugaite and Daniel~M. Roy.
\newblock Computing nonvacuous generalization bounds for deep (stochastic) neural networks with many more parameters than training data.
\newblock In Gal Elidan, Kristian Kersting, and Alexander Ihler, editors, \emph{Proceedings of the Thirty-Third Conference on Uncertainty in Artificial Intelligence, {UAI} 2017, Sydney, Australia, August 11-15, 2017}, 2017.

\bibitem[Germain et~al.(2016)Germain, Bach, Lacoste, and Lacoste-Julien]{germain2016pac}
Pascal Germain, Francis Bach, Alexandre Lacoste, and Simon Lacoste-Julien.
\newblock {PAC}-{Bayesian theory meets Bayesian inference}.
\newblock \emph{Advances in Neural Information Processing Systems}, 2016.

\bibitem[Goodfellow et~al.(2015)Goodfellow, Shlens, and Szegedy]{goodfellow2014explaining}
Ian~J. Goodfellow, Jonathon Shlens, and Christian Szegedy.
\newblock Explaining and harnessing adversarial examples.
\newblock In Yoshua Bengio and Yann LeCun, editors, \emph{International Conference on Learning Representations, {ICLR}}, 2015.

\bibitem[Grosse et~al.(2018)Grosse, Pfaff, Smith, and Backes]{grosse2018limitations}
Kathrin Grosse, David Pfaff, Michael~Thomas Smith, and Michael Backes.
\newblock The limitations of model uncertainty in adversarial settings.
\newblock \emph{arXiv preprint arXiv:1812.02606}, 2018.

\bibitem[Hern{\'a}ndez-Lobato et~al.(2011)Hern{\'a}ndez-Lobato, Hern{\'a}ndez-Lobato, and Dupont]{hernandez2011robust}
Daniel Hern{\'a}ndez-Lobato, Jose Hern{\'a}ndez-Lobato, and Pierre Dupont.
\newblock Robust multi-class gaussian process classification.
\newblock \emph{Advances in neural information processing systems}, 2011.

\bibitem[Hoffman et~al.(2014)Hoffman, Gelman, et~al.]{hoffman2014no}
Matthew~D Hoffman, Andrew Gelman, et~al.
\newblock The no-u-turn sampler: adaptively setting path lengths in hamiltonian monte carlo.
\newblock \emph{Journal of Machine Learning Research}, 2014.

\bibitem[Jin et~al.(2022)Jin, Banerjee, and Mont{\'{u}}far]{jin2022learning}
Hui Jin, Pradeep~Kr. Banerjee, and Guido Mont{\'{u}}far.
\newblock Learning curves for gaussian process regression with power-law priors and targets.
\newblock In \emph{The Tenth International Conference on Learning Representations, {ICLR}}, 2022.

\bibitem[Kim and Ghahramani(2008)]{kim2008outlier}
Hyun-Chul Kim and Zoubin Ghahramani.
\newblock Outlier robust gaussian process classification.
\newblock In \emph{Structural, Syntactic, and Statistical Pattern Recognition: Joint IAPR International Workshop, SSPR \& SPR 2008, Orlando, USA, December 4-6, 2008. Proceedings}. Springer, 2008.

\bibitem[Kirschner et~al.(2020)Kirschner, Bogunovic, Jegelka, and Krause]{kirschner2020distributionally}
Johannes Kirschner, Ilija Bogunovic, Stefanie Jegelka, and Andreas Krause.
\newblock Distributionally robust bayesian optimization.
\newblock In \emph{International Conference on Artificial Intelligence and Statistics}. PMLR, 2020.

\bibitem[Knoblauch et~al.(2022)Knoblauch, Jewson, and Damoulas]{knoblauch2022optimization}
Jeremias Knoblauch, Jack Jewson, and Theodoros Damoulas.
\newblock An optimization-centric view on bayes' rule: Reviewing and generalizing variational inference.
\newblock \emph{Journal of Machine Learning Research}, 2022.

\bibitem[Li et~al.(2023)Li, Wang, Guo, and Wang]{li2023adersarial}
Ang Li, Yifei Wang, Yiwen Guo, and Yisen Wang.
\newblock Adversarial examples are not real features.
\newblock In Alice Oh, Tristan Naumann, Amir Globerson, Kate Saenko, Moritz Hardt, and Sergey Levine, editors, \emph{Neural Information Processing Systems (NeurIPS)}, 2023.

\bibitem[Lotfi et~al.(2022)Lotfi, Finzi, Kapoor, Potapczynski, Goldblum, and Wilson]{lotfi2022pac}
Sanae Lotfi, Marc Finzi, Sanyam Kapoor, Andres Potapczynski, Micah Goldblum, and Andrew~G Wilson.
\newblock {PAC-Bayes} compression bounds so tight that they can explain generalization.
\newblock \emph{Advances in Neural Information Processing Systems}, 2022.

\bibitem[Madry et~al.(2018)Madry, Makelov, Schmidt, Tsipras, and Vladu]{madry2018towards}
Aleksander Madry, Aleksandar Makelov, Ludwig Schmidt, Dimitris Tsipras, and Adrian Vladu.
\newblock Towards deep learning models resistant to adversarial attacks.
\newblock In \emph{International Conference on Learning Representations, {ICLR}}, 2018.

\bibitem[Markelle et~al.()Markelle, Rachel, and Kolby]{uci}
Kelly Markelle, Longjohn Rachel, and Nottingham Kolby.
\newblock The uci machine learning repository.
\newblock URL \url{https://archive.ics.uci.edu}.

\bibitem[McAllester(1998)]{mcallester1998some}
David~A McAllester.
\newblock Some {PAC-Bayesian} theorems.
\newblock In \emph{Proceedings of the eleventh annual conference on Computational learning theory}, 1998.

\bibitem[McCullagh and Nelder(1989)]{mccullagh1989generalized}
P.~McCullagh and J.A. Nelder.
\newblock \emph{Generalized Linear Models, Second Edition}.
\newblock Monographs on Statistics and Applied Probability Series. Chapman \& Hall, 1989.

\bibitem[Nash et~al.(1994)Nash, Sellers, Talbot, Cawthorn, and Ford]{abalone}
Warwick Nash, Tracy Sellers, Simon Talbot, Andrew Cawthorn, and Wes Ford.
\newblock {Abalone}.
\newblock UCI Machine Learning Repository, 1994.

\bibitem[Nielsen(2021)]{nielsen2021geodesic}
Frank Nielsen.
\newblock On geodesic triangles with right angles in a dually flat space.
\newblock In \emph{Progress in Information Geometry: Theory and Applications}, pages 153--190. Springer, 2021.

\bibitem[Opper and Vivarelli(1998)]{opper1998general}
Manfred Opper and Francesco Vivarelli.
\newblock General bounds on bayes errors for regression with gaussian processes.
\newblock \emph{Advances in Neural Information Processing Systems}, 1998.

\bibitem[Ovadia et~al.(2019)Ovadia, Fertig, Ren, Nado, Sculley, Nowozin, Dillon, Lakshminarayanan, and Snoek]{ovadia2019can}
Yaniv Ovadia, Emily Fertig, Jie Ren, Zachary Nado, David Sculley, Sebastian Nowozin, Joshua Dillon, Balaji Lakshminarayanan, and Jasper Snoek.
\newblock Can you trust your model's uncertainty? evaluating predictive uncertainty under dataset shift.
\newblock \emph{Advances in neural information processing systems}, 2019.

\bibitem[Paszke et~al.(2017)Paszke, Gross, Chintala, Chanan, Yang, DeVito, Lin, Desmaison, Antiga, and Lerer]{paszke2017automatic}
Adam Paszke, Sam Gross, Soumith Chintala, Gregory Chanan, Edward Yang, Zachary DeVito, Zeming Lin, Alban Desmaison, Luca Antiga, and Adam Lerer.
\newblock Automatic differentiation in {PyTorch}.
\newblock In \emph{NIPS-W}, 2017.

\bibitem[Patane et~al.(2022)Patane, Blaas, Laurenti, Cardelli, Roberts, and Kwiatkowska]{patane2022adversarial}
Andrea Patane, Arno Blaas, Luca Laurenti, Luca Cardelli, Stephen Roberts, and Marta Kwiatkowska.
\newblock Adversarial robustness guarantees for gaussian processes.
\newblock \emph{Journal of Machine Learning Research}, 2022.

\bibitem[Pedregosa et~al.(2011)Pedregosa, Varoquaux, Gramfort, Michel, Thirion, Grisel, Blondel, Prettenhofer, Weiss, Dubourg, VanderPlas, Passos, Cournapeau, Brucher, Perrot, and Duchesnay]{pedregosa2011cali}
Fabian Pedregosa, Ga{\"{e}}l Varoquaux, Alexandre Gramfort, Vincent Michel, Bertrand Thirion, Olivier Grisel, Mathieu Blondel, Peter Prettenhofer, Ron Weiss, Vincent Dubourg, Jake VanderPlas, Alexandre Passos, David Cournapeau, Matthieu Brucher, Matthieu Perrot, and Edouard Duchesnay.
\newblock Scikit-learn: Machine learning in python.
\newblock \emph{Journal of Machine Learning Research}, 2011.

\bibitem[Quinlan(1993)]{auto_mpg}
R.~Quinlan.
\newblock {Auto MPG}.
\newblock UCI Machine Learning Repository, 1993.

\bibitem[Reeb et~al.(2018)Reeb, Doerr, Gerwinn, and Rakitsch]{reeb2018learning}
David Reeb, Andreas Doerr, Sebastian Gerwinn, and Barbara Rakitsch.
\newblock Learning gaussian processes by minimizing {PAC-Bayesian} generalization bounds.
\newblock \emph{Advances in Neural Information Processing Systems}, 2018.

\bibitem[Savvides et~al.(2024)Savvides, Luu, and Puolam{\"a}ki]{savvides2024error}
Rafael Savvides, Hoang Phuc~Hau Luu, and Kai Puolam{\"a}ki.
\newblock Error bounds for any regression model using gaussian processes with gradient information.
\newblock In \emph{International Conference on Artificial Intelligence and Statistics}. PMLR, 2024.

\bibitem[Seeger(2002)]{seeger2002pac}
Matthias Seeger.
\newblock {PAC-Bayesian} generalisation error bounds for gaussian process classification.
\newblock \emph{Journal of machine learning research}, 2002.

\bibitem[Shafahi et~al.(2019)Shafahi, Huang, Studer, Feizi, and Goldstein]{shhafahi2019are}
Ali Shafahi, W.~Ronny Huang, Christoph Studer, Soheil Feizi, and Tom Goldstein.
\newblock Are adversarial examples inevitable?
\newblock In \emph{International Conference on Learning Representations, {ICLR}}, 2019.

\bibitem[Shalev-Shwartz and Ben-David(2014)]{shalev2014understanding}
Shai Shalev-Shwartz and Shai Ben-David.
\newblock \emph{Understanding Machine Learning: From Theory to Algorithms}.
\newblock Cambridge University Press, 2014.

\bibitem[Sollich(1998)]{sollich1998learning}
Peter Sollich.
\newblock Learning curves for gaussian processes.
\newblock \emph{Advances in neural information processing systems}, 1998.

\bibitem[Sollich(2001)]{sollich2001gaussian}
Peter Sollich.
\newblock Gaussian process regression with mismatched models.
\newblock \emph{Advances in Neural Information Processing Systems}, 2001.

\bibitem[Sollich and Halees(2002)]{sollich2002learning}
Peter Sollich and Anason~S. Halees.
\newblock Learning curves for gaussian process regression: Approximations and bounds.
\newblock \emph{Neural Computation}, 2002.

\bibitem[Suzuki(2012)]{suzuki2012pac}
Taiji Suzuki.
\newblock {PAC-Bayesian} bound for gaussian process regression and multiple kernel additive model.
\newblock In \emph{Conference on Learning Theory}. JMLR Workshop and Conference Proceedings, 2012.

\bibitem[Szegedy et~al.(2014)Szegedy, Zaremba, Sutskever, Bruna, Erhan, Goodfellow, and Fergus]{szegedy2013intriguing}
Christian Szegedy, Wojciech Zaremba, Ilya Sutskever, Joan Bruna, Dumitru Erhan, Ian~J. Goodfellow, and Rob Fergus.
\newblock Intriguing properties of neural networks.
\newblock In \emph{International Conference on Learning Representations, {ICLR}}, 2014.

\bibitem[Tsanas and Xifara(2012)]{energy_efficiency}
Athanasios Tsanas and Angeliki Xifara.
\newblock {Energy Efficiency}.
\newblock UCI Machine Learning Repository, 2012.

\bibitem[Tsipras et~al.(2019)Tsipras, Santurkar, Engstrom, Turner, and Madry]{tsipras2019robustness}
Dimitris Tsipras, Shibani Santurkar, Logan Engstrom, Alexander Turner, and Aleksander Madry.
\newblock Robustness may be at odds with accuracy.
\newblock In \emph{International Conference on Learning Representations}, 2019.

\bibitem[Vanschoren et~al.(2013)Vanschoren, van Rijn, Bischl, and Torgo]{OpenML2013}
Joaquin Vanschoren, Jan~N. van Rijn, Bernd Bischl, and Luis Torgo.
\newblock Openml: Networked science in machine learning.
\newblock \emph{SIGKDD Explorations}, 2013.

\bibitem[Vapnik and Chervonenkis(1971)]{vapnik1971uniform}
VN~Vapnik and A~Ya Chervonenkis.
\newblock On the uniform convergence of relative frequencies of events to their probabilities.
\newblock \emph{Theory of Probability and its Applications}, 1971.

\bibitem[Vito(2008)]{air_quality}
Saverio Vito.
\newblock {Air Quality}.
\newblock UCI Machine Learning Repository, 2008.

\bibitem[Wainwright et~al.(2008)Wainwright, Jordan, et~al.]{wainwright2008graphical}
Martin~J Wainwright, Michael~I Jordan, et~al.
\newblock Graphical models, exponential families, and variational inference.
\newblock \emph{Foundations and Trends{\textregistered} in Machine Learning}, 1\penalty0 (1--2):\penalty0 1--305, 2008.

\bibitem[Wang et~al.(2023)Wang, Ding, Levinboim, Chen, and Soricut]{wang2023improving}
Zifan Wang, Nan Ding, Tomer Levinboim, Xi~Chen, and Radu Soricut.
\newblock Improving robust generalization by direct {PAC-Bayesian} bound minimization.
\newblock In \emph{Proceedings of the IEEE/CVF Conference on Computer Vision and Pattern Recognition}, 2023.

\bibitem[Wicker et~al.(2021)Wicker, Laurenti, Patane, Chen, Zhang, and Kwiatkowska]{wicker2021bayesian}
Matthew Wicker, Luca Laurenti, Andrea Patane, Zhuotong Chen, Zheng Zhang, and Marta Kwiatkowska.
\newblock Bayesian inference with certifiable adversarial robustness.
\newblock In \emph{International Conference on Artificial Intelligence and Statistics}, pages 2431--2439. PMLR, 2021.

\bibitem[Williams and Vivarelli(2000)]{williams2000upper}
Christopher K.~I. Williams and Francesco Vivarelli.
\newblock Upper and lower bounds on the learning curve for gaussian processes.
\newblock \emph{Machine Learning}, 2000.

\bibitem[Zhang(2006)]{zhang2006information}
Tong Zhang.
\newblock Information-theoretic upper and lower bounds for statistical estimation.
\newblock \emph{IEEE Transactions on Information Theory}, 2006.

\end{thebibliography}

\newpage

\onecolumn

\title{Generalization Certificates for Adversarially Robust Bayesian Linear Regression\\(Supplementary Material)}

\section*{Generalization Certificates for Adversarially Robust Bayesian Linear Regression (Supplementary Material)}

\appendix
\section{Adversarial robust loss}
\label{app:adv_loss}
\subsection{The robust posterior}
The variational form of exact Bayes inference can be obtained by minimising the KL divergence of the notional posterior $q'$ from the true posterior $q$ over all probability density functions, as follows.
\begin{align*}
    &\phantom{{}={}} q(\theta \mid \mathcal{D}) \\
    &= \argmin_{q' \in \Pi} KL(q' \Vert q) \\
    &=  \argmin_{q' \in \Pi}\expectation{q'(\theta \mid \mathcal{D})} {\log \frac{q'(\theta \mid \mathcal{D}) \int p(Y \mid X, \theta') \pi(\theta') d\theta' }{ p(Y \mid X, \theta) \pi(\theta)}} \\
    &=\argmin_{q' \in \Pi}\mathop{\mathbb{E}}\limits_{q'(\theta \mid \mathcal{D})}\Bigg[\sum_{i=1}^n  -\log \frac{p(y_i \mid x_i, \theta)}{\underbrace{\int p(y_i \mid x_i, \theta) dy_i}_{=1}}\Bigg]  +  \underbrace{\log \int \prod_{i=1}^n p(y_i \mid x_i, \theta') \pi(\theta') d\theta'}_{\text{const. w.r.t. $q'$}} + KL(q' \Vert \pi ) \numberthis \label{eq:standard_bayes_choices} \\
    &= \argmin_{q' \in \Pi}\expectation{q'(\theta \mid \mathcal{D})}{\sum_{i=1}^n  -\log p(y_i \mid x_i, \theta)} + KL(q' \Vert \pi ) .
\end{align*}
\subsection{Other notions of robust posterior}
Note that in~\eqref{eq:standard_bayes_choices}, we explicitly convey that they likelihood is a proper likelihood (i.e. integrates to $1$) and the posterior is a proper posterior, possessing a normalizing constant which is independent of the functional variable $q'$. 
Thus, from an optimisation-centric view, both terms can be ignored.
When we generalize to Gibbs Bayes posteriors by changing the negative log likelihood to an adversarial loss, we therefore obtain
\begin{align*}
    &\phantom{{}={}} q(\theta \mid \mathcal{D}) =\argmin_{q' \in \Pi}\expectation{q'(\theta \mid \mathcal{D})}{\sum_{i=1}^n  \max_{\Vert \widetilde{x}_i - x_i \Vert_2 \leq \delta }-\log p(y_i \mid x_i, \theta)} + KL(q' \Vert \pi ) = \frac{\exp\big( - \mathcal{L}( \theta, \mathcal{D}) \big) \pi(\theta) }{\int \exp\big( - \mathcal{L}( \theta', \mathcal{D}) \big) \pi(\theta') d\theta' }, \numberthis \label{eq:our_choice}
\end{align*}
with $\mathcal{L}(\theta, \mathcal{D}) = \sum_{i=1}^n  \max_{\Vert \widetilde{x}_i - x_i \Vert_2 \leq \delta }-\log p(y_i \mid x_i, \theta)$.

We note however that this is not the only natural choice of ``robustifying'' posterior inference. 
In particular, starting from~\eqref{eq:standard_bayes_choices}, there are four choices depending on whether we ignore or do not ignore the normalising constants of the likelihood and the posterior when inserting the operator $\max_{\Vert \widetilde{x}_i - x_i \Vert_2 \leq \delta }$.
These four choices all lead to the same standard Bayesian posterior (i.e. $\delta = 0$), but lead to different notions of robust posterior.
In addition to these four choices, one may also consider robustifying the likelihood before the normalizing step, leading to a true likelihood (and thus, standard Bayesian inference)
Our choice~\eqref{eq:our_choice} allows for a tractable loss term, leads to a satisfying theory, and good empirical performance. 
Other choices do not immediately lead to a tractable loss term, and we leave this and investigation of their theory and empirical performance for future work.

\subsection{Proof of \lowercase{\Cref{lm:adv_loss_closed_form_gaussian,lm:adv_loss_closed_form}}}
\begin{proof}
We begin with the Gaussian case, then consider the more general exponential family case, and then return to the Gaussian case.
Choosing a Gaussian likelihood and a linear predictor, up to some constant,
\begin{align*}
    \max_{\widetilde{x}_i: \Vert x_i - \widetilde{x}_i \Vert \leq \delta}-\log p(y_i \mid \widetilde{x}_i, \theta) &= \max_{\widetilde{x}_i: \Vert x_i - \widetilde{x}_i \Vert \leq \delta} (y_i - \widetilde{x}_i^\top \theta)^2 \\
    &= \max_{\widetilde{x}_i: \Vert x_i - \widetilde{x}_i \Vert \leq \delta} \big(y_i -x_i^\top \theta - \theta^\top (\widetilde{x}_i -x_i) \big)^2 \\
    &= \max_{\widetilde{x}_i: \Vert x_i - \widetilde{x}_i \Vert \leq \delta} \big(y_i -x_i^\top \theta - \Vert \theta \Vert_2  \delta \cos \gamma \big)^2,
\end{align*}
where $\gamma$ is the angle between $\theta$ and $\widetilde{x}_i - x_i$.
The argument of $(\cdot)^2$ is maximally positive or negative when $\cos\gamma$ is $\pm 1$ and shares the same sign as $y_i - x_i^\top \theta$. Thus
\begin{align*}
    \max_{\widetilde{x}_i: \Vert x_i - \widetilde{x}_i \Vert \leq \delta}-\log p(y_i \mid \widetilde{x}_i, \theta) &= \big( |y_i - x_i^\top \theta| + \Vert \theta \Vert \delta\big)^2 \\
    &= (y_i - x_i^\top \theta)^2 + 2 \delta \Vert \theta \Vert |y_i - x_i^\top \theta|  + \Vert \theta \Vert^2 \delta^2,
\end{align*}
That is, $\tilde{x}_i =  \delta \sign(x_i^\top \theta - y_i)\frac{\theta}{\|\theta\|} + x_i$. 

Consider the Bregman divergence and apply the law of cosines (e.g. Property 1 of~\citet{nielsen2021geodesic}),
\begin{align*}
 \max_{\widetilde{x}: \Vert x - \widetilde{x} \Vert \leq \delta} d_\phi \big( \theta^\top \widetilde{x}, y^\ast \big) &= \max_{\widetilde{x}: \Vert x - \widetilde{x} \Vert \leq \delta} d_\phi \big( \theta^\top \widetilde{x}, \theta^\top x \big) + d_\phi \big( \theta^\top x, y^\ast \big) - \theta^\top (\widetilde{x} - x)\big(\nabla \phi(y^\ast) - \nabla \phi(\theta^\top x) \big) \\
 &= \max_{\widetilde{x}: \Vert x - \widetilde{x} \Vert \leq \delta} \phi(\theta^\top \widetilde{x}) - \phi(\theta^\top x) - \nabla \phi(y^\ast) \theta^\top(\widetilde{x} - x) + d_\phi(\theta^\top x, y^\ast).
\end{align*}
This is a convex objective on a convex constraint set $\Vert x - \widetilde{x} \Vert_2^2 \leq \delta^2$, so there exists a unique maximum on an extremal point on the constraint set.
The KKT conditions give that at the optimal,
\begin{align*}
    \big(\nabla(\phi(\theta^\top \widetilde{x}) - \nabla \phi(y^\ast)\big) \theta - 2 \lambda (\widetilde{x} - x) = 0,
\end{align*}
for Lagrange multiplier $\lambda \leq 0$.
Therefore, $\widetilde{x}$ satisfies the implicit equation
\begin{align*}
    \widetilde{x} = \frac{\big(\nabla(\phi(\theta^\top \widetilde{x}) - \nabla\phi(y^\ast)\big) \theta}{2 \lambda} + x. \numberthis \label{eq:implicit_xtilde}
\end{align*}
We must have the solution on the extremal, so 
\begin{align*}
    \delta = \Big| \frac{\big(\nabla(\phi(\theta^\top \widetilde{x}) - \nabla\phi(y^\ast)\big) }{2 \lambda} \Big| \Vert \theta \Vert_2 \quad \text{and so} \quad \lambda = \Big| \frac{\nabla(\phi(\theta^\top \widetilde{x}) - \nabla\phi(y^\ast) }{2 \delta} \Big| \Vert \theta \Vert_2.
\end{align*}
Plugging $\lambda$ back into~\eqref{eq:implicit_xtilde}, we find that the optimal $\widetilde{x}$ is a linear combination of $\theta$ and $x$,
\begin{align*}
    \widetilde{x} = \delta \underbrace{\sign \big(\nabla(\phi(\theta^\top \widetilde{x}) - \nabla\phi(y^\ast) \big)}_{:= s \in \{-1, 1\}} \Vert \theta \Vert_2^{-1} \theta + x.
\end{align*}
The maximum value is then
\begin{align*}
    \phi(s \delta \Vert \theta \Vert_2 + \theta^\top x) - \phi(\theta^\top x) - \nabla \phi(y^\ast) s \delta \Vert \theta \Vert_2 + d_\phi(\theta^\top x, y^\ast).
\end{align*}
Finally, note that $\nabla \phi(y^\ast) = \nabla \phi\big( (\nabla \phi)^{-1}(y)\big)=y$.
We may then compute the maximum by testing the two $s \in \{ -1, 1\}$, and choosing the value of $s$ which gives the maximum result.
In the case of Gaussian loss (i.e. squared error), the parameters are self-dual and we have
\begin{align*}
    &\phantom{{}={}} \phi(s \delta \Vert \theta \Vert_2 + \theta^\top x) - \phi(\theta^\top x) - \nabla \phi(y) s \delta \Vert \theta \Vert_2 + d_\phi(\theta^\top x, y) \\
    &=\delta^2 \Vert \theta \Vert_2^2 + 2 s \delta \Vert \theta \Vert_2 (\theta^\top x) - 2 y s \delta \Vert \theta \Vert_2 + \Vert y - \theta^\top x \Vert_2^2 \\
    &= \delta^2 \Vert \theta \Vert_2^2 + 2 s \delta \Vert \theta \Vert_2  \big( (\theta^\top x) -  y  \big) + \Vert y - \theta^\top x \Vert_2^2,
\end{align*}
the maxima being $\delta^2 \Vert \theta \Vert_2^2 + 2  \delta \Vert \theta \Vert_2  \big| (\theta^\top x) -  y  \big| + \Vert y - \theta^\top x \Vert_2^2$ with $s = \sign (\theta^\top x - y)$.
\end{proof}

\clearpage

\section{Proof of CGF bounds for standard and adversarial NLL losses in \MakeLowercase{\Cref{lm:cgf_bounds_std_adv_loss}}}
\label{app:cgf_std_adv_loss}

We first derive the following helpful \Cref{lm:adv_loss_bounds,lm:std_gaussian_int} to derive the CGF bounds and the generalization certificates.

\begin{lemma}[Upper and lower bounds for adversarial NLL loss]
\label{lm:adv_loss_bounds} 
Using the closed-form of the adversarial NLL loss, the upper and lower bounds are
\begin{align*}
     \ell_\delta(\theta, \mathcal{D}) &= \sum_{i=1}^n \lp  \frac{1}{2\sigma^2}\big(|y_i - x_i^\top \theta| + \Vert \theta \Vert \delta\big)^2 + \frac{1}{2} \log \lp 2\pi \sigma^2\rp \rp \\
    &\leq \sum_{i=1}^n \lp \frac{1}{2\sigma^2} \Big( 2(y_i - x_i^\top \theta)^2 + 2\Vert \theta \Vert^2 \delta^2 \Big) + \frac{1}{2} \log \lp 2\pi \sigma^2\rp \rp \quad ; (a-b)^2 \geq 0 \implies a^2+b^2 \geq 2ab \\
     \ell_\delta(\theta, \mathcal{D}) &\geq \sum_{i=1}^n \lp  \frac{1}{2\sigma^2} \Big( (y_i - x_i^\top \theta)^2 + \Vert \theta \Vert^2 \delta^2 \Big) + \frac{1}{2} \log \lp 2\pi \sigma^2\rp \rp
\end{align*}
\qed
\end{lemma}

\begin{lemma}[Standard Gaussian integral] \label{lm:std_gaussian_int} The integral of the form 
$\int \exp\left(-\theta^\top M \theta + 2b^\top \theta\right) d\theta$ evaluates to $\sqrt{\frac{\pi^d}{\det M}} \exp\left(b^\top M^{-1}b\right).$

\begin{proof}
\begin{align*}
    &-\theta^\top M \theta + 2b^\top \theta = -(\theta - M^{-1}b)^\top M (\theta - M^{-1}b) + b^\top M^{-1}b \quad; \text{Completing the square} \\
    &\text{Therefore, integral becomes } \int \exp\left(-(\theta - M^{-1}b)^\top M (\theta - M^{-1}b) + b^\top M^{-1}b\right) d\theta \\
    &= \exp\left(b^\top M^{-1}b\right) \int \exp\left(-(\theta - M^{-1}b)^\top M (\theta - M^{-1}b)\right) d\theta \\
    &= \exp\left(b^\top M^{-1}b\right) \sqrt{\frac{\pi^d}{\det M}} \quad; \int \exp\left(-\phi^\top M \phi\right) d\phi = \sqrt{\frac{\pi^d}{\det M}}
\end{align*}
\end{proof}
\end{lemma}

\paragraph{Proof of CGF bounds for standard NLL loss}
Now we derive the CGF bound for standard NLL loss in the following. Note that similar derivation is done in \citet{germain2016pac}.
\begin{proof}
\begin{align*}
    &\log \mathbb{E}_{\theta}\mathbb{E}_{x_i}\expectation{y_i \mid x_i}{\exp \lp t \lp \mathbb{E}_{x_i}\expectation{y_i \mid x_i}{\frac{1}{2\sigma^2}(y_i - x_i^\top \theta)^2} - \frac{1}{2\sigma^2}(y_i - x_i^\top \theta)^2 \rp \rp} \\
    &= \log \mathbb{E}_{\theta}\mathbb{E}_{x_i}\expectation{y_i \mid x_i}{\exp \lp \frac{t}{2\sigma^2} \lp \mathbb{E}_{x_i}\expectation{y_i \mid x_i}{(y_i - x_i^\top \theta)^2}\rp \exp \lp - \frac{t}{2\sigma^2}(y_i - x_i^\top \theta)^2 \rp \rp} \quad; t>0 \implies \exp(-t(.)^2) \leq 1 \\
    &\leq \log \mathbb{E}_{\theta}\mathbb{E}_{x_i}\expectation{y_i \mid x_i}{\exp \lp \frac{t}{2\sigma^2} \lp \mathbb{E}_{x_i}\expectation{y_i \mid x_i}{(y_i - x_i^\top \theta)^2}\rp \rp} \\
    &= \log \mathbb{E}_{\theta}\mathbb{E}_{x_i}\expectation{\epsilon_i}{\exp \lp \frac{t}{2\sigma^2} \lp \mathbb{E}_{x_i}\expectation{\epsilon_i}{(x_i^\top (\theta^* -\theta) + \epsilon_i)^2}\rp \rp}  \\
    &= \log \expectation{\theta}{\exp \lp \frac{t}{2\sigma^2} \lp \sigma_x^2 \| \theta^* -\theta \|^2 + \sigma^2  \rp\rp} \quad; \epsilon_i \sim \mathcal{N}(0, \sigma^2), \mathbb{E}[x_i] = 0, \mathbb{E}[\|x_i\|^2] = \sigma_x^2 \\
    &= \log \int \exp \lp \frac{t}{2\sigma^2}\lp \sigma_x^2 \theta^\top \theta - 2\sigma_x^2 \theta^{*\top} \theta +  \sigma_x^2 \|\theta^*\|^2 +  \sigma^2 \rp - \frac{1}{2\sigma_p^2}\theta^\top \theta  \rp \frac{1}{\sqrt{2\pi \sigma_p^2}^d} d\theta \quad; \theta \sim \mathcal{N}(0, \sigma^2_p I) \\
    &= \log \int \exp \lp - \lp \frac{1-t\sigma_p^2 \sigma_x^2/\sigma^2}{2\sigma^2_p} \rp \theta^\top \theta - \frac{t\sigma_x^2}{\sigma^2} \theta^{*\top} \theta + \frac{t\sigma_x^2}{2\sigma^2}  \|\theta^*\|^2 + \frac{t}{2}  \rp \frac{1}{\sqrt{2\pi \sigma_p^2}^d} d\theta 
\end{align*}
\begin{align*}
    &= \log \sqrt{\frac{\pi 2 \sigma^2_p}{1-t\sigma^2_p \sigma_x^2/\sigma^2}}^d \exp \lp \frac{t^2 \sigma_x^4 \|\theta^*\|^2\sigma^2_p/2\sigma^2}{1-t\sigma_p^2 \sigma_x^2/\sigma^2} + \frac{t \sigma_x^2}{2\sigma^2} \|\theta^*\|^2 + \frac{t}{2} \rp \frac{1}{\sqrt{2\pi \sigma_p^2}^d} \quad; t< \frac{\sigma^2}{\sigma^2_p \sigma_x^2},  \text{Lemma \ref{lm:std_gaussian_int}} \\
    &= \frac{d}{2}\log \frac{1}{1-t\sigma^2_p \sigma_x^2/\sigma^2} + \frac{t \sigma_x^2 \|\theta^*\|^2/2\sigma^2}{1-t\sigma_p^2 \sigma_x^2/\sigma^2} + \frac{t}{2} \\
    &\leq \frac{t\sigma^2_p \sigma_x^2d/\sigma^2}{2\lp 1-t\sigma^2_p \sigma_x^2/\sigma^2\rp} + \frac{t \sigma_x^2 \|\theta^*\|^2/\sigma^2}{2 \lp 1-t\sigma_p^2 \sigma_x^2/\sigma^2 \rp} + \frac{t}{2} \quad ; -\log (1-x) \leq \frac{x}{1-x} \\
    &= \frac{t^2s^2}{2(1-tc)}
\end{align*}
From above we get $s^2 = \frac{1}{t} \lp \frac{\sigma^2_p \sigma_x^2d}{\sigma^2} + \frac{\sigma_x^2 \|\theta^*\|^2}{\sigma^2} + \lp 1-\frac{t\sigma^2_p \sigma_x^2}{\sigma^2}\rp \rp $, $c=\frac{\sigma^2_p \sigma_x^2}{\sigma^2}$ and $t \in (0,\frac{\sigma^2}{\sigma^2_p \sigma_x^2})$.
\end{proof}

\paragraph{Proof of CGF bounds for adversarial NLL loss}
The cumulant generating function of the adversarial loss can be bounded similar to standard loss as follows. 
\begin{proof}
\begin{align*}
    &\log \expectation{\theta}{\exp \lp \frac{t}{2\sigma^2} \mathbb{E}_{x_i}\expectation{y_i \mid x_i}{ (y_i - x_i^\top \theta)^2 + 2 \delta \Vert \theta \Vert |y_i - x_i^\top \theta|  + \Vert \theta \Vert^2 \delta^2} \rp} \\
    &\leq \log \expectation{\theta}{\exp \lp \frac{t}{2\sigma^2} \mathbb{E}_{x_i}\expectation{y_i \mid x_i}{ 2(y_i - x_i^\top \theta)^2 + 2 \Vert \theta \Vert^2 \delta^2} \rp} \quad; \text{Lemma \ref{lm:adv_loss_bounds}} \\
    &= \log \expectation{\theta}{\exp \lp \frac{t}{2\sigma^2} \mathbb{E}_{x_i}\expectation{\epsilon_i}{ 2(x_i^\top \lp \theta^* - \theta \rp + \epsilon_i)^2 + 2\Vert \theta \Vert^2 \delta^2} \rp} \\
    &= \log \expectation{\theta}{\exp \lp \frac{t}{\sigma^2} \sigma_x^2 \| \theta^* - \theta \|^2 + t + \frac{t}{\sigma^2}\Vert \theta \Vert^2 \delta^2 \rp} \\
    &= \log \int \exp\lp \lp \frac{t\sigma_x^2}{\sigma^2} + \frac{t\delta^2}{\sigma^2} - \frac{1}{2\sigma^2_p} \rp \|\theta\|^2 - \frac{2t \sigma_x^2}{\sigma^2} \theta^{*\top} \theta + \frac{t \sigma_x^2}{\sigma^2} \|\theta^*\|^2 + t  \rp \frac{1}{\sqrt{2\pi \sigma^2_p}^d} d\theta \\
    &= \log \sqrt{\frac{\pi 2 \sigma^2_p}{1-2\sigma^2_pt\lp \sigma_x^2+\delta^2 \rp/\sigma^2}}^d \exp \lp \frac{t^2\sigma_x^4\|\theta^{*}\|^2 2\sigma^2_p/\sigma^2}{1-2\sigma^2_pt\lp \sigma_x^2+\delta^2 \rp/\sigma^2}+ \frac{t \sigma_x^2}{\sigma^2} \|\theta^*\|^2 + t \rp \frac{1}{\sqrt{2\pi \sigma^2_p}^d} \\
    &=\frac{d}{2}\log \frac{1}{1-2\sigma^2_pt\lp \sigma_x^2+\delta^2 \rp/\sigma^2} + \frac{\frac{t \sigma_x^2}{\sigma^2} \|\theta^*\|^2 + t \lp 1-\frac{2\sigma^2_p t}{\sigma^2}\lp \sigma_x^2+\delta^2 \rp \rp}{1-2\sigma^2_pt\lp \sigma_x^2+\delta^2 \rp/\sigma^2} \\
    &\leq \frac{\sigma_p^2td \lp \sigma_x^2 + \delta^2 \rp/\sigma^2}{1-2\sigma^2_pt\lp \sigma_x^2+\delta^2 \rp/\sigma^2}+ \frac{\frac{t \sigma_x^2}{\sigma^2} \|\theta^*\|^2 + t \lp 1-\frac{2\sigma^2_p t}{\sigma^2}\lp \sigma_x^2+\delta^2 \rp \rp}{1-2\sigma^2_pt\lp \sigma_x^2+\delta^2 \rp/\sigma^2} \\
    &= \frac{t^2s^2}{2(1-tc)}
\end{align*}
For adversarial loss, 
$s^2 = \frac{2}{t} \lp cd + \frac{\sigma_x^2 \|\theta^*\|^2}{\sigma^2} + \lp 1-ct \rp \rp$, $c=\frac{2\sigma^2_p\lp \sigma_x^2+\delta^2 \rp}{\sigma^2}$ and $t\in \lp 0, \frac{\sigma^2}{2\sigma^2_p\lp \sigma_x^2+\delta^2 \rp} \rp$.
\end{proof}

\section{Proof of Theorems in \MakeLowercase{\Cref{ss:robust_cert}}}
\label{app:pac_bounds_proof_robust_posterior}

In this section, we derive \Cref{thm:std_post_std_loss,thm:std_post_adv_loss,thm:adv_post_std_loss,thm:adv_post_adv_loss,thm:adv_post_adv_loss_gen} using \Cref{th:pac_bayes_bounded_cgf}. From the PAC-Bayesian bounds for bounded CGF loss theorem, it is clear that we need to bound the expected training risk plus the KL divergence between the posterior and prior. In the bound derivation, we require to bound the negative log normalizing constants of the posteriors which will be presented first in \Cref{lm:neg_log_z_bayes,lm:neg_log_z_robust}. 

\begin{lemma}[Negative log normalizing constant of the Bayes posterior] 
The normalizing constant $z$ of Bayes posterior $q$ is 
\begin{align*}
    {z} &= \int \exp\Big( -\mathcal{L}(\theta, \mathcal{D}) \Big) \pi(\theta) d\theta \\
    &= \int \exp\Big( -\frac{1}{2\sigma^2}\sum_{i=1}^n (y_i - x_i^\top \theta)^2 \Big) \pi(\theta) d\theta \\
    &= \frac{\exp(-Y^\top Y/2\sigma^2)}{\sqrt{2\pi\sigma^2_p}^d}\int \exp\Big( -\theta^\top \lp \frac{1}{2\sigma^2}X^\top X + \frac{1}{2\sigma^2_p}I \rp \theta + \frac{1}{2\sigma^2}2 Y^\top X \theta \Big) \Big)  d\theta  \\
    &= \frac{\exp \lp -\frac{1}{2\sigma^2}Y^\top Y + \frac{1}{2\sigma^2}Y^\top X \lp \frac{1}{\sigma^2}X^\top X + \frac{1}{\sigma^2_p}I \rp^{-1} X^\top Y\frac{1}{\sigma^2} \rp}{\sqrt{ \det \lp \frac{\sigma_p^2}{\sigma^2} X^\top X +  I\rp}}
\end{align*}
\begin{align*}
     \log \frac{1}{z} &= \frac{1}{2}\log \det \lp \frac{\sigma_p^2}{\sigma^2} X^\top X + I \rp + \frac{1}{2\sigma^2} \lp Y^\top Y - Y^\top X \lp \frac{1}{\sigma^2} X^\top X + I \rp^{-1} X^\top Y\frac{1}{\sigma^2} \rp  \\
    &=   \frac{1}{2}\log  \det \lp \frac{\sigma_p^2}{\sigma^2} X^\top X + I \rp + \frac{1}{2\sigma^2} Y^\top \lp I + \frac{\sigma_p^2}{\sigma^2}  XX^\top  \rp^{-1} Y \quad; \text{Woodbury Matrix Identity}
\end{align*}
\qed
\label{lm:neg_log_z_bayes}
\end{lemma}

Similar to above \Cref{lm:neg_log_z_bayes}, we derive the negative log normalizing constant of the robust posterior in the following.

\begin{lemma}[Negative log normalizing constant of the robust posterior]
The normalizing constant $z_\delta$ of robust posterior $q_\delta$ is
\begin{align*}
    z_\delta &= \int \exp\Big( -\mathcal{L}_\delta(\theta, \mathcal{D}) \Big) \pi(\theta) d\theta \\
    &\geq \int \exp\Big( -{\frac{1}{\sigma^2}} \sum_{i=1}^n \Big( (y_i - x_i^\top \theta)^2 + \Vert \theta \Vert^2 \delta^2 \Big) \Big) \pi(\theta) d\theta \\
    &= \frac{1}{\sqrt{2\pi\sigma^2_p}^d}\int \exp\Bigg( - \Big( \frac{n \delta^2}{{\sigma^2}}\Vert \theta \Vert^2  + \frac{1}{2\sigma^2_p}\Vert \theta \Vert^2 + \frac{1}{{\sigma^2}} \Vert Y - X\theta \Vert^2  \Big) \Bigg)  d\theta  \quad; \pi(\theta) \sim \mathcal{N}(0,\sigma^2_p I)\\
    &= \frac{1}{\sqrt{2\pi\sigma^2_p}^d}\int \exp\Bigg( - \Big( \frac{ n\delta^2}{{\sigma^2}} \theta^\top \theta  + \frac{1}{2\sigma^2_p}\theta^\top \theta + {\frac{1}{\sigma^2}}\lp \Vert Y \Vert^2 - 2 Y^\top X \theta + \theta^\top X^\top X \theta \rp \Big) \Bigg)  d\theta  \\
    &= \frac{1}{\sqrt{2\pi\sigma^2_p}^d} \exp(-\frac{1}{\sigma^2}\Vert Y\Vert^2) \int \exp\Bigg( - \theta^\top \big( (\frac{n\delta^2}{{\sigma^2}} + \frac{1}{2\sigma^2_p})I + \frac{X^\top X}{{\sigma^2}}  \big)\theta + \frac{2 Y^\top X\theta}{{\sigma^2}} \Bigg)  d\theta \quad; \text{Lemma \ref{lm:std_gaussian_int}}  \\
    &=  \sqrt{\frac{1}{\det \lp 2{\frac{\sigma_p^2}{\sigma^2}}X^\top X + ({\frac{\sigma_p^2}{\sigma^2}}2 n\delta^2 + 1)I\rp}} \exp(-\frac{\Vert Y\Vert^2}{{\sigma^2}}) \exp\lp \frac{2Y^\top X}{{\sigma^4}} \lp {\frac{2}{\sigma^2}}X^\top X + ({\frac{2 n\delta^2}{\sigma^2}} + \frac{1}{\sigma^2_p})I\rp^{-1}X^\top Y\rp 
\end{align*}
\begin{align*}
    \log \frac{1}{z_\delta}&\leq  
    \frac{1}{2} \log \det \lp 2{\frac{\sigma_p^2}{\sigma^2}}X^\top X + ({\frac{\sigma_p^2}{\sigma^2}}2 n\delta^2 + 1)I\rp +  {\frac{1}{\sigma^2}} Y^\top  \lp I - \frac{2}{{\sigma^2}}X \lp {\frac{2}{\sigma^2}}X^\top X + ({\frac{1}{\sigma^2}}2 n\delta^2 + \frac{1}{\sigma^2_p})I\rp^{-1} X^\top  \rp Y  \\
    &= \frac{1}{2} \log  \det \lp 2{\frac{\sigma_p^2}{\sigma^2}}X^\top X + ({\frac{\sigma_p^2}{\sigma^2}}2 n\delta^2 + 1)I\rp +  {\frac{1}{\sigma^2}} Y^\top  \lp I + \frac{2\sigma^2_p}{2n\delta^2\sigma^2_p + \sigma^2} XX^\top  \rp^{-1} Y  \quad; \text{Woodbury Matrix Identity} \\
    &= \frac{1}{2} \log \det \lp 2{\frac{\sigma_p^2}{\sigma^2}}X^\top X + ({\frac{\sigma_p^2}{\sigma^2}}2 n\delta^2 + 1)I\rp + {\frac{\sigma^2}{\sigma^2 \lp 2n\delta^2 \sigma_p^2 + \sigma^2 \rp}} Y^\top  \lp I \lp \frac{\sigma_p^2}{\sigma^2}2n\delta^2 + 1 \rp + \frac{2\sigma^2_p}{\sigma^2} XX^\top  \rp^{-1} Y
\end{align*}
\qed
\label{lm:neg_log_z_robust}
\end{lemma}

With the above bounds derived, we are now ready to derive the generalization certificate using \Cref{th:pac_bayes_bounded_cgf}. Following holds for all the cases:

$(i)$ The CGF bound of respective loss-NLL in case of standard generalization and adversarial NLL in case of adversarial generalization)-appears in the theorem. This directly implies that the constants $c$ and $s$ should follow \Cref{lm:cgf_bounds_std_adv_loss} according to the considered setting.

$(ii)$ The free parameter in the CGF bound, $t$ is choosen as $t=1$, which is also done in the prior work \citep{germain2016pac}. This means $1 \in \lp 0, 1/c \rp$, that is, $c < 1$.

Consequently, the effective \Cref{th:pac_bayes_bounded_cgf} is stated below for clarity.  

\begin{theorem}[\Cref{th:pac_bayes_bounded_cgf}] \label{th:pac_bayes_bounded_cgf_t1}
Consider data $\mathcal{D}$ and any loss $\ell \big(\theta, (x,y)\big)$ with its corresponding expected and empirical generalization errors $R(\theta)=\expectation{(x,y) \sim \mathcal{P}}{\ell \big(\theta, (x,y)\big)}$ and $r(\theta)=\frac{1}{n} \mathcal{L}(\theta, \mathcal{D})$, respectively.  
Let the CGF of the loss $\psi(t) \leq \frac{t^2s^2}{2 \lp 1-tc \rp}$  be bounded, where for some constant $c \in \lp 0,. 1 \rp$ and $t \in (0,1/c)$. 
Then, we have, with probability at least $1-\beta$, for all densities 
$\rho(\theta)$,
    $$\expectation{\theta \sim \rho}{R(\theta)}
\leq 
\expectation{\theta \sim \rho}{r(\theta)} + \frac{1}{n} \mathrm{KL}({\rho} \| \pi) + \frac{1}{n} \log \frac{1}{\beta}
+ \frac{s^2}{2 \lp 1-c \rp}.$$
\end{theorem}

Therefore, we only need to bound the following for $\rho$ being standard Bayes $q$ and robust posterior $q_\delta$, and $\ell$ being NLL and adversarial NLL.
\begin{align}
\expectation{\theta \sim \rho}{r(\theta)} + \frac{1}{n} \mathrm{KL}({\rho} \| \pi). \label{eq:eff_pac_bound}
\end{align}

\subsection{Standard generalization of Bayes posterior (\MakeLowercase{\Cref{thm:std_post_std_loss})}}
\label{pf:std_post_std_loss}
In this case, $\rho = q$ and $\ell$ is NLL in \Cref{eq:eff_pac_bound} which reduces the expression to the following.
\begin{align*}
    \expectation{\theta \sim q}{r(\theta)} + \frac{1}{n} \mathrm{KL}({q} \| \pi) 
    &= \frac{1}{n} \int \frac{1}{{z}}\ell(\theta, \mathcal{D}) \exp(-\ell(\theta, \mathcal{D}))\pi(\theta) d\theta + \frac{1}{n} \int \frac{1}{{z}} \exp(-\ell(\theta, \mathcal{D}))\pi(\theta) \log \lp \frac{\exp(-\ell(\theta, \mathcal{D}))\pi(\theta)}{z\pi(\theta)} \rp d\theta \\
    &= \frac{1}{n} \int \frac{1}{z} \exp(-\ell(\theta, \mathcal{D}))\pi(\theta) \log \lp \frac{1}{z} \rp d\theta \\
    &= \frac{1}{n} \log \frac{1}{z} \stepcounter{equation}\tag{\theequation}\label{eq:std_post_std_loss_normz}
\end{align*}

Therefore, we obtain \Cref{thm:std_post_std_loss} by substituting \Cref{lm:neg_log_z_bayes} in \Cref{eq:std_post_std_loss_normz} and combining it with the \Cref{th:pac_bayes_bounded_cgf_t1}.
\qed

\subsection{Adversarial generalization of robust posterior $\delta=\deltat$ (\MakeLowercase{\Cref{thm:adv_post_adv_loss})}}

Notice that the above derivation in \Cref{pf:std_post_std_loss} holds for $\rho=q_\delta$ and $\ell$ is adversarial loss with $\delta$ perturbation with $z_\delta$ as the normalization constant. That is, the perturbation at train and test-time are the same $\delta=\deltat$. 

\begin{align*}
    \expectation{\theta \sim q_\delta}{r(\theta)} + \frac{1}{n} \mathrm{KL}({q_\delta} \| \pi) 
    &= \frac{1}{n} \log \frac{1}{z_\delta} \stepcounter{equation}\tag{\theequation}\label{eq:adv_post_adv_loss_normz}
\end{align*}

Therefore, we obtain \Cref{thm:adv_post_adv_loss} by substituting \Cref{lm:neg_log_z_robust} in \Cref{eq:adv_post_adv_loss_normz} and combining it with the \Cref{th:pac_bayes_bounded_cgf_t1}.
\qed

\subsection{Adversarial generalization of Bayes posterior (\MakeLowercase{\Cref{thm:std_post_adv_loss})}}

In this case, $\rho = q$ and $\ell$ is adversarial NLL with $\deltat$ perturbation in \Cref{eq:eff_pac_bound}. We upper bound it as follows.

For ease of notation, only in the following, we use short hand notations for $\ell$ to denote $\ell(\theta, \mathcal{D})$ and $\ell_{\deltat}$ to denote $\ell_{\deltat}(\theta, \mathcal{D})$.

\begin{align*}
    \expectation{\theta \sim {{q}}}{\frac{1}{n} \ell_{\deltat} (\theta, \mathcal{D})} + \frac{1}{n} \mathrm{KL}({{q}} \| \pi) 
    &= \frac{1}{n} \int \frac{1}{{z}}\ell_{\deltat}\exp(-{\ell})\pi(\theta) d\theta + \frac{1}{n} \int \frac{1}{{z}} \exp(-{\ell})\pi(\theta) \log \lp \frac{\exp(-{\ell})\pi(\theta)}{{z}\pi(\theta)} \rp d\theta \\
    &= \frac{1}{n} \int \frac{1}{{z}}\lp \ell_{\deltat} - \ell \rp\exp(-{\ell})\pi(\theta) d\theta + \frac{1}{n} \int \frac{1}{{z}} \exp(-{\ell})\pi(\theta) \log \lp \frac{1}{{z}} \rp d\theta \\
    &\leq \frac{1}{n} \int \frac{1}{{z}}\lp 2\ell +  {\frac{1}{2\sigma^2}} 2n \|\theta\|^2 \delta^2 - \ell \rp\exp(-{\ell})\pi(\theta) d\theta + \frac{1}{n} \log \frac{1}{{z}} \quad; \text{Lemma \ref{lm:adv_loss_bounds}} \\
    &\leq \frac{1}{n} \log \int \frac{1}{{z}} \exp \lp \ell +  {\frac{1}{2\sigma^2}}2n \|\theta\|^2 \delta^2 \rp \exp(-{\ell})\pi(\theta) d\theta +  \frac{1}{n} \log \frac{1}{{z}} \quad; \text{Jensen's ineq.} \\
    &= \frac{1}{n} \log \frac{1}{{z}} \int  \exp \lp \lp \frac{2n \delta^2}{ {2\sigma^2}} - \frac{1}{2\sigma_p^2} \rp \|\theta\|^2  \rp \frac{1}{\sqrt{2\pi \sigma_p^2}^d} d\theta +  \frac{1}{n} \log \frac{1}{{z}} \\
    &= \frac{1}{n} \log \frac{1}{{z}} \sqrt{\frac{2\pi \sigma_p^2 {\sigma^2}}{ {\sigma^2}- {2}n\delta^2 \sigma_p^2}}^d \frac{1}{\sqrt{2\pi \sigma_p^2}^d} +   \frac{1}{n} \log \frac{1}{{z}} \\
    &=  \frac{2}{n} \log \frac{1}{{z}} + \frac{d}{2n} \log \frac{1}{1- {\frac{2}{\sigma^2}}n\delta^2 \sigma_p^2} \\
    &\leq  \frac{2}{n} \log \frac{1}{{z}} + \frac{d}{2n} \frac{ {\frac{2}{\sigma^2}}n\delta^2 \sigma_p^2}{1- {\frac{2}{\sigma^2}}n\delta^2 \sigma_p^2} \quad; -\log(1-x) \leq \frac{x}{1-x} \\
    &= \frac{2}{n} \log \frac{1}{{z}} + \frac{d\delta^2 \sigma_p^2 {/\sigma^2}}{1-2n\delta^2 \sigma_p^2 {/\sigma^2}} \stepcounter{equation}\tag{\theequation}\label{eq:std_post_adv_loss_normz}
\end{align*}

We obtain \Cref{thm:std_post_adv_loss} by substituting \Cref{lm:neg_log_z_bayes} in \Cref{eq:std_post_adv_loss_normz} and combining it with the \Cref{th:pac_bayes_bounded_cgf_t1}.
\qed

\subsection{Standard generalization of robust posterior (\MakeLowercase{\Cref{thm:adv_post_std_loss})}}

In this case, $\rho = q_\delta$ and $\ell$ is NLL in \Cref{eq:eff_pac_bound}. We upper bound it as follows.

For ease of notation, only in the following, we use short hand notations for $\ell$ to denote $\ell(\theta, \mathcal{D})$ and $\ell_{\delta}$ to denote $\ell_{\delta}(\theta, \mathcal{D})$.

\begin{align*}
    \expectation{\theta \sim { q_\delta }}{\frac{1}{n}{\ell}(\theta, \mathcal{D})} + \frac{1}{n} \mathrm{KL}({ q_\delta } \| \pi) 
    &= \frac{1}{n} \int \frac{1}{ z_\delta }{\ell}\exp(-\ell_{\delta})\pi(\theta) d\theta + \frac{1}{n} \int \frac{1}{ z_\delta } \exp(-\ell_{\delta})\pi(\theta) \log \lp \frac{\exp(-\ell_{\delta})\pi(\theta)}{ z_\delta \pi(\theta)} \rp d\theta \\
    &= \frac{1}{n} \int \frac{1}{ z_\delta }\lp -\ell_{\delta}+ \ell \rp\exp(-\ell_{\delta})\pi(\theta) d\theta + \frac{1}{n} \int \frac{1}{ z_\delta } \exp(-\ell_{\delta})\pi(\theta) \log \lp \frac{1}{ z_\delta } \rp d\theta \\
    &\leq \frac{1}{n} \int \frac{1}{ z_\delta }\lp -\ell - {\frac{1}{2\sigma^2}}n \|\theta\|^2 \delta^2 + \ell \rp\exp(-\ell_{\delta})\pi(\theta) d\theta + \frac{1}{n} \log \frac{1}{ z_\delta } \quad; \text{Lemma \ref{lm:adv_loss_bounds}} \\
    &\leq \frac{1}{n} \log \int \frac{1}{ z_\delta } \exp \lp -{\frac{1}{2\sigma^2}}n \|\theta\|^2 \delta^2 \rp \exp(-\ell_{\delta})\pi(\theta) d\theta +  \frac{1}{n} \log \frac{1}{ z_\delta } \quad; \text{Jensen's ineq.} \\
    &= \frac{1}{n} \log \int  \exp \lp \lp -{\frac{1}{2\sigma^2}}2n \delta^2 - \frac{1}{2\sigma_p^2} \rp \|\theta\|^2 - \ell \rp \frac{1}{\sqrt{2\pi \sigma_p^2}^d} d\theta  +  \frac{2}{n} \log \frac{1}{ z_\delta } \stepcounter{equation}\tag{\theequation}\label{eq:adv_post_std_loss_normz}
\end{align*}

The evaluation of the above integral $\int  \exp \lp \lp -{\frac{1}{2\sigma^2}}2n \delta^2 - \frac{1}{2\sigma_p^2} \rp \|\theta\|^2 - \ell \rp \frac{1}{\sqrt{2\pi \sigma_p^2}^d} d\theta  +  \frac{2}{n} \log \frac{1}{ z_\delta }$ is as follows:
\begin{align*}
    &= \frac{1}{\sqrt{2\pi\sigma^2_p}^d}\int \exp\Bigg( -\frac{1}{2} \Big( {\frac{1}{2\sigma^2}}4 n \Vert \theta \Vert^2 \delta^2 + \frac{1}{\sigma^2_p}\Vert \theta \Vert^2 + {\frac{2}{2\sigma^2}} \Vert Y - X\theta \Vert^2  \Big) \Bigg)  d\theta  \quad; \pi(\theta) \sim \mathcal{N}(0,\sigma^2_p I)\\
    &= \frac{1}{\sqrt{2\pi\sigma^2_p}^d}\int \exp\Bigg( -\frac{1}{2} \Big( {\frac{1}{2\sigma^2}}4 n\delta^2 \theta^\top \theta  + \frac{1}{\sigma^2_p}\theta^\top \theta + {\frac{1}{2\sigma^2}} \lp 2 \Vert Y \Vert^2 - 4 Y^\top X \theta + 2 \theta^\top X^\top X \theta \rp \Big) \Bigg)  d\theta  \\
    &= \frac{1}{\sqrt{2\pi\sigma^2_p}^d} \exp(-{\frac{1}{2\sigma^2}}\Vert Y\Vert^2) \int \exp\Bigg( - \theta^\top \big( ({\frac{1}{2\sigma^2}}2 n\delta^2 + \frac{1}{2\sigma^2_p})I + {\frac{1}{2\sigma^2}} X^\top X  \big)\theta + {\frac{2}{2\sigma^2}} Y^\top X\theta \Bigg)  d\theta \quad; \text{Lemma \ref{lm:std_gaussian_int}}  \\
    &=  \sqrt{\frac{1}{\det \lp  {\frac{\sigma_p^2}{\sigma^2}}X^\top X + ( {\frac{\sigma_p^2}{\sigma^2}}2 n\delta^2 + 1)I\rp}} \exp(- {\frac{1}{2\sigma^2}}\Vert Y\Vert^2) \exp\lp  {\frac{1}{2\sigma^4}} Y^TX \lp  {\frac{1}{\sigma^2}} X^\top X + ( {\frac{1}{\sigma^2}}2 n\delta^2 + \frac{1}{\sigma^2_p})I\rp^{-1}X^TY\rp \\
    &=  \sqrt{\frac{1}{\det \lp  {\frac{\sigma_p^2}{\sigma^2}}X^\top X + ( {\frac{\sigma_p^2}{\sigma^2}}2 n\delta^2 + 1)I\rp}} \exp  {\frac{1}{2\sigma^2}} \lp - Y^T \lp I -  {\frac{1}{\sigma^2}}X \lp  {\frac{1}{\sigma^2}}X^\top X + ( {\frac{1}{\sigma^2}}2 n\delta^2 + \frac{1}{\sigma^2_p})I\rp^{-1}X^T\rp Y \rp \\
     &=  \sqrt{\frac{1}{\det \lp  {\frac{\sigma_p^2}{\sigma^2}}X^\top X + ( {\frac{\sigma_p^2}{\sigma^2}}2 n\delta^2 + 1)I\rp}} \exp \lp -  {\frac{1}{2\sigma^2}}Y^T \lp I +  {\frac{1}{\sigma^2}}XX^T \lp \frac{\sigma_p^2\sigma^2}{\sigma^2+2 n \delta^2 \sigma^2_p}\rp\rp^{-1} Y \rp \\
     \stepcounter{equation}\tag{\theequation}\label{eq:adv_post_std_loss_normz_2}
\end{align*}
Therefore, we obtain \Cref{thm:adv_post_std_loss} by substituting \Cref{eq:adv_post_std_loss_normz_2} in \Cref{eq:adv_post_std_loss_normz} and combining it with the \Cref{th:pac_bayes_bounded_cgf_t1}.
\qed

\clearpage
\subsection{Adversarial generalization of robust posterior when $\delta \ne \deltat$ (\MakeLowercase{\Cref{thm:adv_post_adv_loss_gen})}}

In this case, $\rho = q_\delta$ and $\ell$ is adversarial NLL with $\deltat$ perturbation in \Cref{eq:eff_pac_bound}. We upper bound it as follows.
For ease of notation, only in the following, we use short hand notations for $\ell_\delta$ to denote $\ell_\delta(\theta, \mathcal{D})$ and $\ell_{\deltat}$ to denote $\ell_{\deltat}(\theta, \mathcal{D})$.

\begin{align*}
    \expectation{\theta \sim {q_\delta}}{\frac{1}{n}{\ell}_{\deltat}(\theta, \mathcal{D})} + \frac{1}{n} \mathrm{KL}({q_\delta} \| \pi) 
    &= \frac{1}{n} \int \frac{1}{z_\delta }{\ell}_{\deltat}\exp(-{\ell}_\delta)\pi(\theta) d\theta + \frac{1}{n} \int \frac{1}{z_\delta } \exp(-{\ell}_\delta)\pi(\theta) \log \lp \frac{\exp(-{\ell}_\delta)\pi(\theta)}{z_\delta \pi(\theta)} \rp d\theta \\
    &= \frac{1}{n} \int \frac{1}{z_\delta }\lp {\ell}_{\deltat} - {\ell}_\delta \rp\exp(-{\ell}_\delta)\pi(\theta) d\theta + \frac{1}{n} \int \frac{1}{z_\delta } \exp(-{\ell}_\delta)\pi(\theta) \log \lp \frac{1}{z_\delta } \rp d\theta \\
    &\leq  \frac{1}{n} \int \frac{1}{z_\delta }\lp 2\ell + \frac{2n \| \theta \|^2 \deltat^2}{2\sigma^2} - \ell - \frac{n\|\theta \|^2\delta^2}{2\sigma^2} \rp\exp(-{\ell}_\delta)\pi(\theta) d\theta + \frac{1}{n}  \log \frac{1}{z_\delta } \quad; \text{Lemma \ref{lm:adv_loss_bounds}} \\
    &\leq \frac{1}{n} \log \int \frac{1}{z_\delta } \exp \lp \ell + \frac{n \|\theta\|^2 \lp 2\deltat^2 - \delta^2\rp}{2\sigma^2} \rp \exp(-{\ell}_\delta)\pi(\theta) d\theta +  \frac{1}{n} \log \frac{1}{z_\delta } \quad; \text{Jensen's ineq.} \\
    &= \frac{1}{n} \log \frac{1}{z_\delta } \int  \exp \lp \lp \frac{n \lp 2\deltat^2 - 2\delta^2 \rp}{2\sigma^2} - \frac{1}{2\sigma_p^2} \rp \|\theta\|^2  \rp \frac{1}{\sqrt{2\pi \sigma_p^2}^d} d\theta +  \frac{1}{n} \log \frac{1}{z_\delta } \\
    &= \frac{1}{n} \log \frac{1}{z_\delta } \sqrt{\frac{2\pi \sigma_p^2}{1-2n\lp \deltat^2 -\delta^2 \rp \sigma_p^2/\sigma^2}}^d \frac{1}{\sqrt{2\pi \sigma_p^2}^d} +   \frac{1}{n} \log \frac{1}{z_\delta } \\
    &=  \frac{2}{n} \log \frac{1}{z_\delta } + \frac{d}{2n} \log \frac{1}{1-2n\lp \deltat^2 -\delta^2 \rp \sigma_p^2/\sigma^2} \\
    &\leq  \frac{2}{n} \log \frac{1}{z_\delta } + \frac{d}{2n} \frac{2n\lp \deltat^2 -\delta^2 \rp \sigma_p^2/\sigma^2}{1-2n\lp \deltat^2 -\delta^2 \rp \sigma_p^2/\sigma^2} \quad; -\log(1-x) \leq \frac{x}{1-x} \\
    &= \frac{2}{n} \log \frac{1}{z_\delta } + \frac{\lp \deltat^2 -\delta^2 \rp \sigma_p^2 d/\sigma^2}{1-2n\lp \deltat^2 -\delta^2 \rp \sigma_p^2/\sigma^2} \stepcounter{equation}\tag{\theequation}\label{eq:adv_post_adv_loss_normz_gen}
\end{align*}

Substituting \Cref{lm:neg_log_z_robust} in \Cref{eq:adv_post_adv_loss_normz_gen} and combining it with the \Cref{th:pac_bayes_bounded_cgf_t1} proves \Cref{thm:adv_post_adv_loss_gen}.

\FloatBarrier
\section{Datasets and implementation details}
\label{app:exp}

The real datasets used in the experiments are given in \Cref{tab:data_stat}, and they are available in UCI repository \citep{uci} or OpenML \citep{OpenML2013}.
We use it 70-30 train-test split to learn and test the posteriors.
The code is developed in Pytorch \citep{paszke2017automatic} and use Pyro package \citep{bingham2019pyro} for NUTS distribution sampler. 
All the experiments are run on CPU of Apple M1 chip with 16GB memory. The run time is between seconds upto a few minutes.

\begin{table}[h]
    \centering
    \begin{tabular}{lcc}
\toprule
Dataset            & Number of samples         & Data dimension \\ \midrule
Abalone            & 4177 & 10             \\
Air Foil           & 1503                      & 5              \\ 
Air Quality        & 7355                      & 11             \\ 
Auto MPG           & 393                       & 9              \\ 
California Housing & 20640                     & 8              \\ 
Energy Efficiency  & 768                       & 8              \\ 
Wine Quality       & 1599                      & 11             \\ \bottomrule
\end{tabular}
    \caption{Real datasets}
    \vspace{-0.4cm}
    \label{tab:data_stat}
\end{table}

\clearpage

\section{Additional experimental results on real data}
\label{app:exp_real}

We provide the additional results on the real datasets for prior variance $\sigma_p^2 = \frac{1}{100}$ in \Cref{tab:real_data_p100} and $\frac{1}{9}$ in \Cref{tab:real_data_p9}. The experimental results are consistent with the results in \Cref{tab:real_data}. We observe that informed choice of prior favors robust posterior in terms of adversarial generalization.

\begin{table*}[h]
    \centering
    \begin{tabular}{lccccc}
    \toprule
    \multirow{2}{*}{Dataset}  
       & \multicolumn{2}{c}{Standard generalization (NLL) $\ell$} & \phantom{}& \multicolumn{2}{c}{Adversarial generalization (adv-NLL) $\ell_{\deltat}$} \\
       & Bayes posterior $q$  & Robust posterior  $q_{\delta}$ & \phantom{}& Bayes posterior $q$ & Robust posterior $q_{\delta}$  \\
       \midrule
       Abalone & \textbf{1.1664} \tiny{$\pm$ 0.014} & 1.1797 \tiny{$\pm$ 0.014} && 1.2221 \tiny{$\pm$ 0.014} & \textbf{1.2172} \tiny{$\pm$ 0.015} \\
       Air Foil & \textbf{1.1714} \tiny{$\pm$ 0.008} & 1.1788 \tiny{$\pm$ 0.009} && \textbf{1.2183} \tiny{$\pm$ 0.009} & {1.2219} \tiny{$\pm$ 0.009} \\
       Air Quality & \textbf{0.9668} \tiny{$\pm$ 0.002} & 0.9674 \tiny{$\pm$ 0.002} && 0.9800 \tiny{$\pm$ 0.002} & \textbf{0.9791} \tiny{$\pm$ 0.002} \\
       Auto MPG & \textbf{1.0295} \tiny{$\pm$ 0.010} & 1.0305 \tiny{$\pm$ 0.010} && \textbf{1.0471} \tiny{$\pm$ 0.011} & {1.0474} \tiny{$\pm$ 0.011} \\
       California Housing & \textbf{1.1195} \tiny{$\pm$ 0.003} & 1.1280 \tiny{$\pm$ 0.005} && 1.1862 \tiny{$\pm$ 0.003} & \textbf{1.1768} \tiny{$\pm$ 0.005} \\
       Energy Efficiency & \textbf{0.9834} \tiny{$\pm$ 0.009} & 0.9838 \tiny{$\pm$ 0.009} && 1.0007 \tiny{$\pm$ 0.010} & \textbf{1.0006} \tiny{$\pm$ 0.010} \\
       Wine Quality & \textbf{1.2323} \tiny{$\pm$ 0.006} & 1.2329 \tiny{$\pm$ 0.006} && 1.2642 \tiny{$\pm$ 0.006} & \textbf{1.2614} \tiny{$\pm$ 0.006} \\
       \bottomrule
    \end{tabular}
    \caption{Test NLL and adversarial NLL of Bayes and robust posteriors on real datasets. The prior variance is set to $\sigma_p^2 = \frac{1}{100}$. The robust posterior is trained with $\delta=0.1$ in the adversarial NLL loss, and adversarial generalization is evaluated using the same training-time perturbation ($\deltat=0.1$). The adversarial generalization results demonstrate that the robust posterior $q_\delta$ is consistently more robust than the Bayes posterior $q$. For both standard and adversarial generalization, the best-performing model for each dataset is highlighted in bold.
    }
    \label{tab:real_data_p100}
\end{table*}

\begin{table*}[h]
    \centering
    \begin{tabular}{lccccc}
    \toprule
    \multirow{2}{*}{Dataset}  
       & \multicolumn{2}{c}{Standard generalization (NLL) $\ell$} & \phantom{}& \multicolumn{2}{c}{Adversarial generalization (adv-NLL) $\ell_{\deltat}$} \\
       & Bayes posterior $q$  & Robust posterior  $q_{\delta}$ & \phantom{}& Bayes posterior $q$ & Robust posterior $q_{\delta}$  \\
       \midrule
       Abalone & \textbf{1.1585} \tiny{$\pm$ 0.013} & 1.1726 \tiny{$\pm$ 0.014} && 1.2554 \tiny{$\pm$ 0.011} & \textbf{1.2176} \tiny{$\pm$ 0.015} \\
       Air Foil & \textbf{1.1657} \tiny{$\pm$ 0.009} & 1.1694 \tiny{$\pm$ 0.009} && 1.2191 \tiny{$\pm$ 0.009} & \textbf{1.2176} \tiny{$\pm$ 0.009} \\
       Air Quality & \textbf{0.9665} \tiny{$\pm$ 0.002} & 0.9670 \tiny{$\pm$ 0.002} && 0.9826 \tiny{$\pm$ 0.003} & \textbf{0.9791} \tiny{$\pm$ 0.002} \\
       Auto MPG & 1.0231 \tiny{$\pm$ 0.006} & \textbf{1.0228} \tiny{$\pm$ 0.007} && 1.0552 \tiny{$\pm$ 0.006} & \textbf{1.0469} \tiny{$\pm$ 0.008} \\
       California Housing & \textbf{1.1193} \tiny{$\pm$ 0.003} & 1.1267 \tiny{$\pm$ 0.005} && 1.1909 \tiny{$\pm$ 0.003} & \textbf{1.1768} \tiny{$\pm$ 0.005} \\
       Energy Efficiency & \textbf{0.9712} \tiny{$\pm$ 0.006} & 0.9733 \tiny{$\pm$ 0.007} && 0.9990 \tiny{$\pm$ 0.007} & \textbf{0.9947} \tiny{$\pm$ 0.008} \\
       Wine Quality & 1.2338 \tiny{$\pm$ 0.006} & \textbf{1.2321} \tiny{$\pm$ 0.006} && 1.2697 \tiny{$\pm$ 0.008} & \textbf{1.2625} \tiny{$\pm$ 0.006} \\
       \bottomrule
    \end{tabular}
    \caption{Test NLL and adversarial NLL of Bayes and robust posteriors on real datasets. The prior variance is set to $\sigma_p^2 = \frac{1}{9}$. The robust posterior is trained with $\delta=0.1$ in the adversarial NLL loss, and adversarial generalization is evaluated using the same training-time perturbation ($\deltat=0.1$). The adversarial generalization results demonstrate that the robust posterior $q_\delta$ is consistently more robust than the Bayes posterior $q$. For both standard and adversarial generalization, the best-performing model for each dataset is highlighted in bold.
    }
    \label{tab:real_data_p9}
\end{table*}

\end{document}